\documentclass[lettersize,journal]{IEEEtran}
\usepackage{amsmath,amsfonts}
\usepackage{algorithmic}
\usepackage{algorithm}
\usepackage{array}
\usepackage[caption=false,font=normalsize,labelfont=sf,textfont=sf]{subfig}
\usepackage{textcomp}
\usepackage{stfloats}
\usepackage{url}
\usepackage{verbatim}
\usepackage{graphicx}
\usepackage{cite}
\usepackage{amssymb}
\usepackage{color}
\usepackage[dvipsnames]{xcolor} 
\usepackage{bbm}
\usepackage{booktabs}
\usepackage{multirow}
\usepackage{multicol}
\usepackage{makecell}
\usepackage[switch]{lineno}

\usepackage{hyperref}
\hypersetup{hidelinks,
	colorlinks=true,
	allcolors=blue,
	pdfstartview=Fit,
	breaklinks=true}

\begin{document}

\title{Fusion-then-Distillation: Toward Cross-modal Positive Distillation for Domain Adaptive 3D Semantic Segmentation}

\author{Yao~Wu,
        Mingwei~Xing,
        Yachao~Zhang,
        Yuan~Xie,~\IEEEmembership{Member,~IEEE},
        Yanyun~Qu,~\IEEEmembership{Member,~IEEE}

\thanks{This work was supported by the National Natural Science Foundation of China under Grant No.62176224, No.62306165; Natural Science Foundation of Chongqing under No.CSTB2023NSCQ-JQX0007; China Computer Federation (CCF)-Lenovo Blue Ocean Research Fund; China Academy of Railway Sciences No.2023YJ357.
}

\thanks{Yao Wu and Yanyun Qu are with the School of Informatics, Xiamen University, Xiamen 361005, China (e-mail: wuyao@stu.xmu.edu.cn, yyqu@xmu.edu.cn).

Mingwei Xing is with the Institute of Artificial Intelligence, Xiamen University, Xiamen 361005, China (e-mail: xingmingwei@stu.xmu.edu.cn).

Yachao Zhang is with the Tsinghua Shenzhen International Graduate School, Tsinghua University, Shenzhen 518071, China (e-mail: yachaozhang@sz.tsinghua.edu.cn).

Yuan Xie is with the School of Computer Science and Technology, East China Normal University, Shanghai 200062, China, and Chongqing Institute of East China Normal University, Chongqing, 401120, China (e-mail: yxie@cs.ecnu.edu.cn).

(Corresponding authors: Yanyun Qu.)

}


}

\markboth{Submitted to TCSVT}%
{Shell \MakeLowercase{\textit{et al.}}: A Sample Article Using IEEEtran.cls for IEEE Journals}


\maketitle

\begin{abstract}
In cross-modal unsupervised domain adaptation, a model trained on source-domain data (\textit{e.g.,} synthetic) is adapted to target-domain data (\textit{e.g.,} real-world) without access to target annotation.
Previous methods seek to mutually mimic cross-modal outputs in each domain, which enforces a class probability distribution that is agreeable in different domains.
However, they overlook the complementarity brought by the heterogeneous fusion in cross-modal learning.
In light of this, we propose a novel fusion-then-distillation (FtD++) method to explore cross-modal positive distillation of the source and target domains for 3D semantic segmentation.
FtD++ realizes distribution consistency between outputs not only for 2D images and 3D point clouds but also for source-domain and augment-domain.
Specially, our method contains three key ingredients. First, we present a model-agnostic feature fusion module to generate the cross-modal fusion representation for establishing a latent space. In this space, two modalities are enforced maximum correlation and complementarity.
Second, the proposed cross-modal positive distillation preserves the complete information of multi-modal input and combines the semantic content of the source domain with the style of the target domain, thereby achieving domain-modality alignment.
Finally, cross-modal debiased pseudo-labeling is devised to model the uncertainty of pseudo-labels via a self-training manner.
Extensive experiments report state-of-the-art results on several domain adaptive scenarios under unsupervised and semi-supervised settings. Code is available at \url{https://github.com/Barcaaaa/FtD-PlusPlus}.
\end{abstract}


\section{Introduction}
\IEEEPARstart{L}{iDAR} sensors yield point clouds that provide a powerful way to perceive, interpret, and reconstruct the complex 3D visual world.
As the core of autonomous driving, efficient and robust 3D scene understanding is a crucial precondition of precise positioning, path planning, and smart transport.
3D semantic segmentation is a key technology in 3D scene understanding, as it can densely assign specific semantic classes to each point.
Like other computer-vision tasks, 3D semantic segmentation faces the domain shift issue, which results in performance degradation on a new unlabeled dataset (target-domain) with a different distribution from the labeled training dataset (source-domain).
For instance, a 3D model learned on synthetic point clouds collected by the Unity game engine usually performs terribly on real point clouds collected by the LiDAR sensor.
Annotating large-scale real datasets for every new scenario is a straightforward solution, but it leans on labor-intensive and time-consuming manual operations, especially for the tasks demanding point-wise annotations.

\begin{figure}[t]
    \centering
    \subfloat[vehicle]{\includegraphics[width=0.46\linewidth]{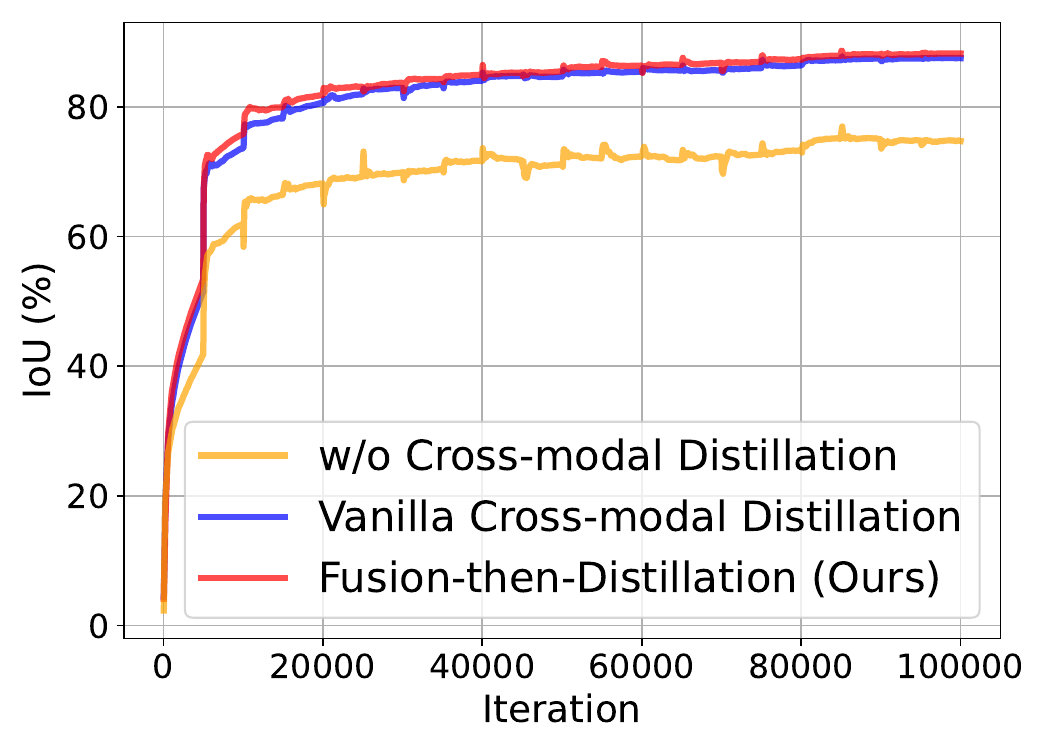}
    \label{fig:cls_fus_0}}
    \hfil
    \subfloat[driv. surf.]{\includegraphics[width=0.46\linewidth]{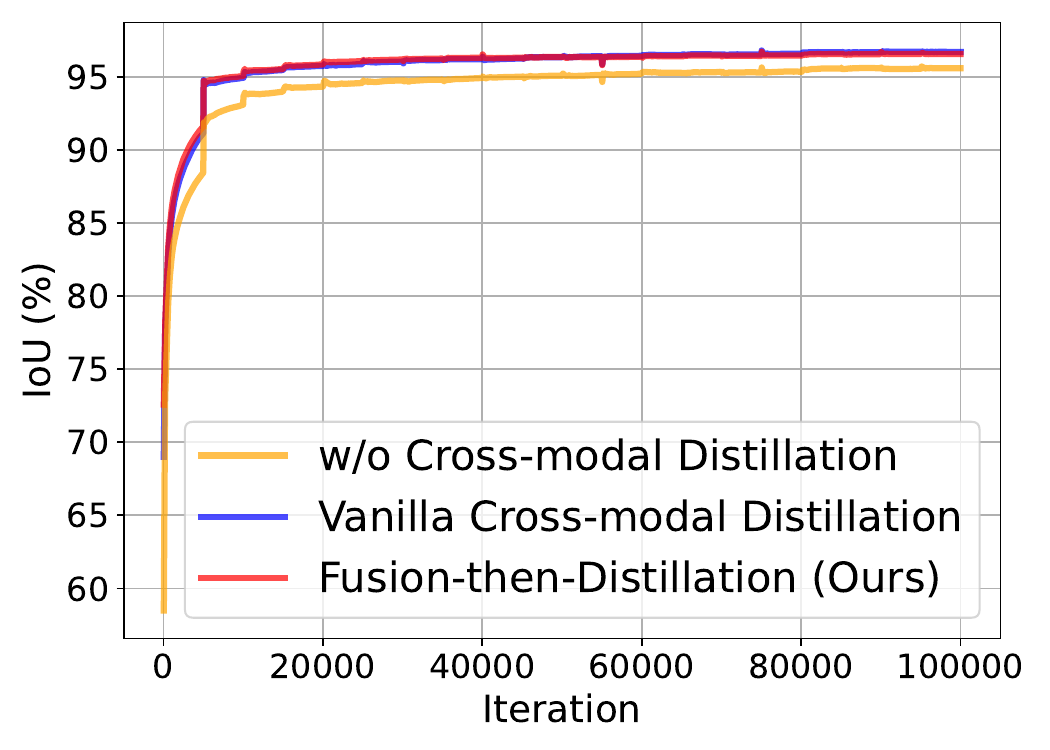}
    \label{fig:cls_fus_1}}
    \hfil
    \subfloat[sidewalk]{\includegraphics[width=0.46\linewidth]{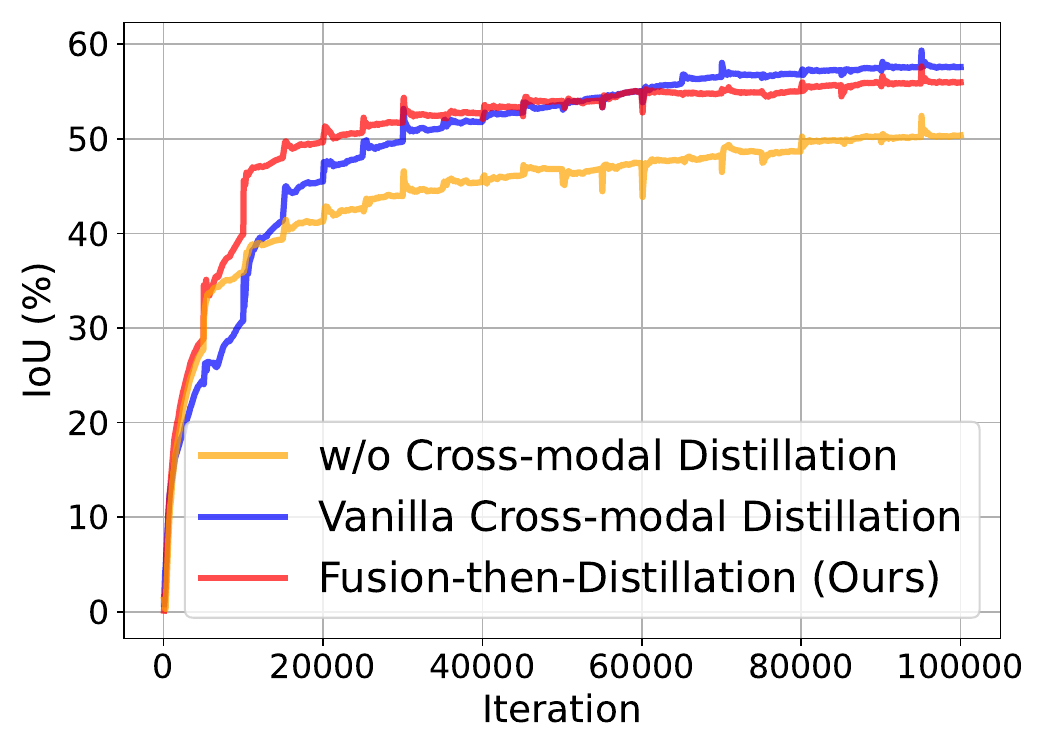}
    \label{fig:cls_fus_2}}
    \hfil
    \subfloat[terrain]{\includegraphics[width=0.46\linewidth]{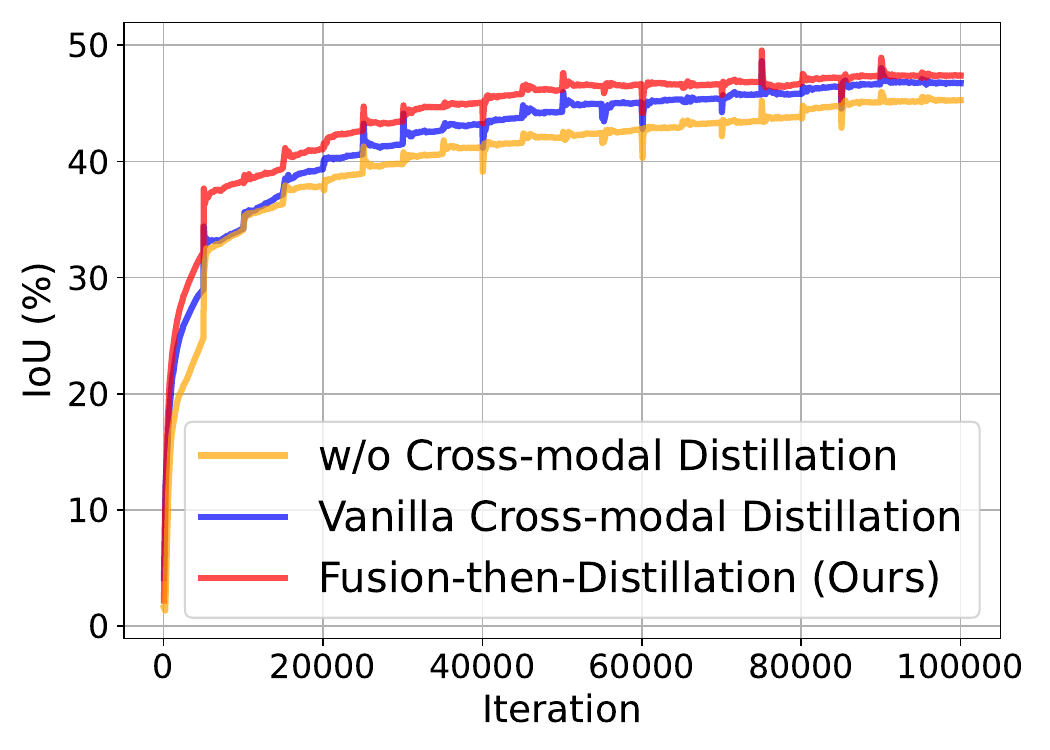}
    \label{fig:cls_fus_3}}
    \hfil
    \subfloat[manmade]{\includegraphics[width=0.46\linewidth]{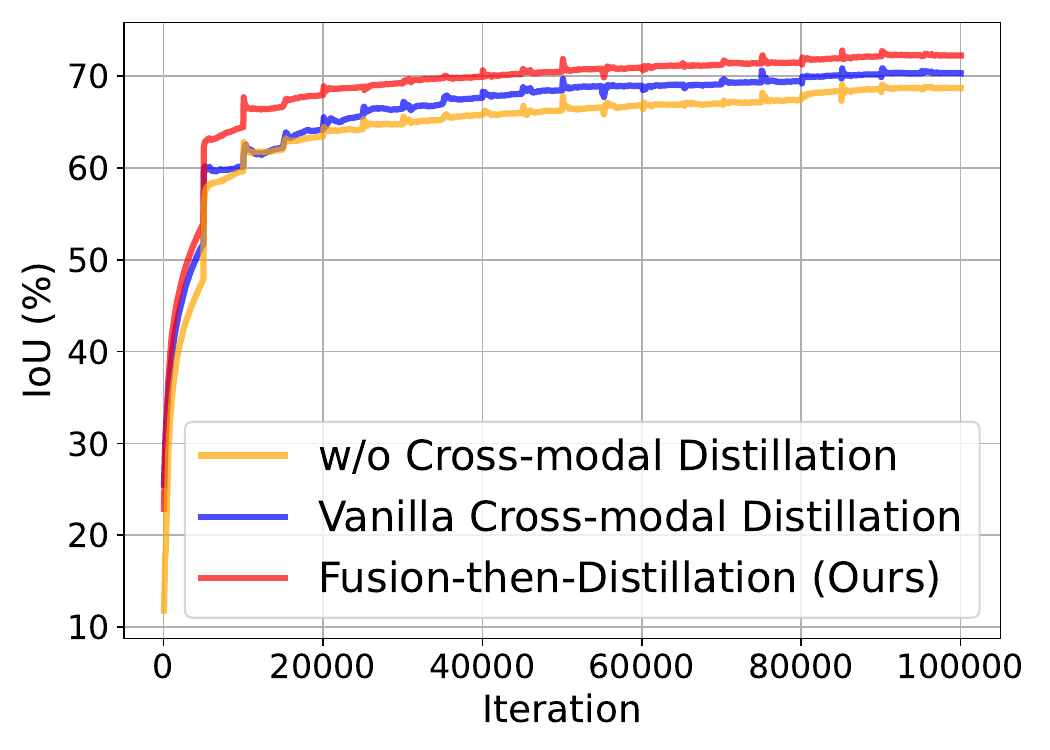}
    \label{fig:cls_fus_4}}
    \hfil
    \subfloat[vegetation]{\includegraphics[width=0.46\linewidth]{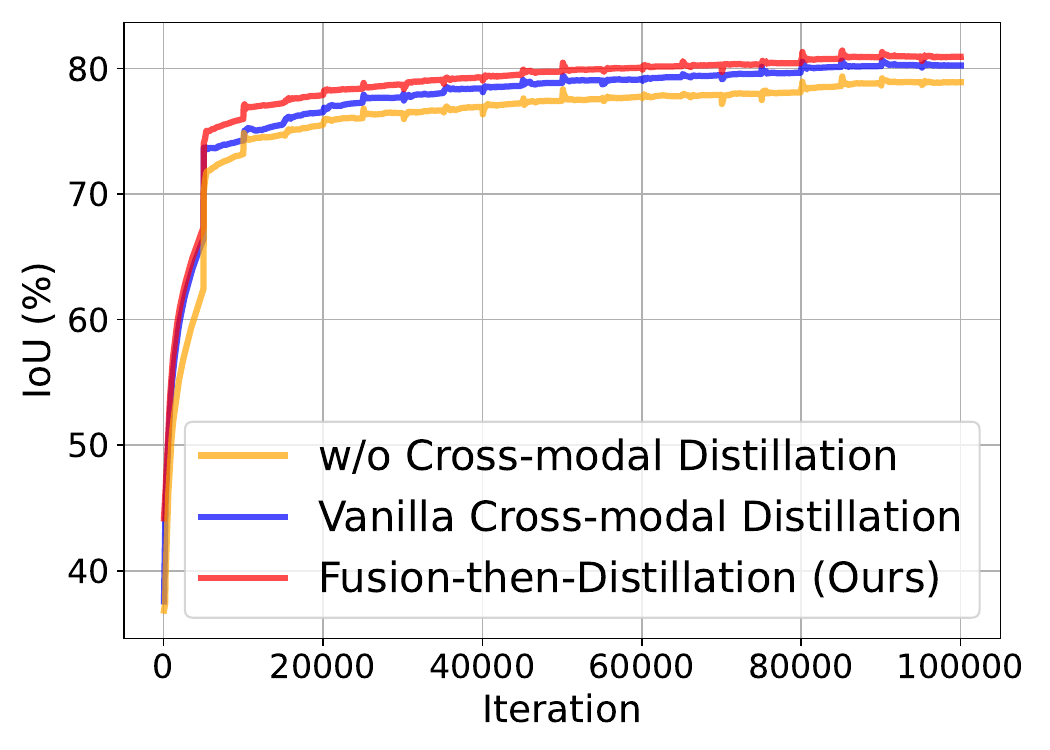}
    \label{fig:cls_fus_5}}
    \caption{Per-class IoU(\%) on ``Day$\to$Night'' scenario by using w/o cross-modal distillation, vanilla cross-modal distillation, and our fusion-then-distillation.}
    \label{fig:per_class_iou}
\end{figure}
\begin{figure*}[t]
    \centering
    \includegraphics[width=0.9\linewidth]{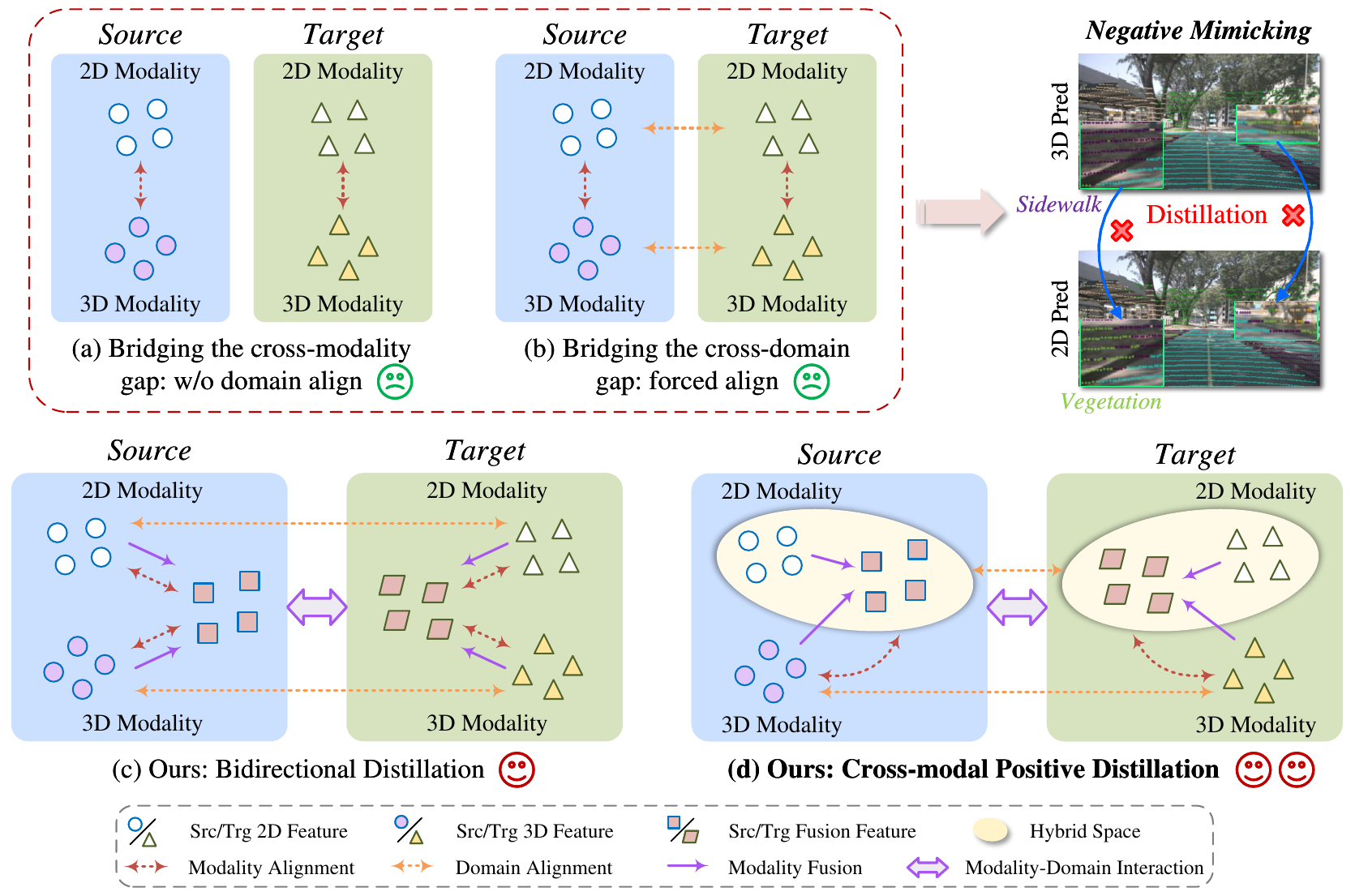}
    \caption{The comparison of four types of cross-modal UDA methods, from top to bottom: (a) Bridging the cross-modality gap; (b) Bridging the cross-domain gap; (c) Bidirectional Distillation~\cite{BFtD-xMUDA} and (d) Cross-modal Positive Distillation proposed in this work. The core idea of (c) and (d) is to explore whether and how cross-modal fusion representation facilitates an efficient and robust UDA model.}
    \label{fig:motivation}
\end{figure*}

The Unsupervised Domain Adaptation (UDA) technique is proposed precisely to alleviate the discrepancy in data distribution between the source and target domains and has made significant progress in both 2D~\cite{ProDA, HRDA, CaoZLCXW23, ZhouFGPCLSM23} and 3D~\cite{CosMix, KongQL23, WangLX23, LiLSLLS23, MicheleBPVMC24} domains.
Recently, cross-modal UDA has attracted considerable attention, reinforcing the contextual information in 2D images by exploiting the geometry of 3D point clouds, and vice versa.
Existing methods are categorized into bridging the cross-modality gap and the cross-domain gap.
The former is tackled by using mutual learning~\cite{xMUDA, MM-TTA, MoPA, wu2024clip2uda} or contrastive learning~\cite{CMCL} (see Fig.~\ref{fig:motivation}(a)).
Building upon them, the latter incorporates domain alignment by learning domain-invariant features~\cite{AUDA, Dual-Cross} and modality-specific information to complement each other~\cite{DsCML, SSE-xMUDA} (see Fig.~\ref{fig:motivation}(b)).
Overall, these methods all rely on modality complementation as a core foundation. As shown in Fig.~\ref{fig:per_class_iou}, we evaluate the predictions for each class in the target training data both with and without cross-modal distillation. It is observed that all classes achieve performance gains under the impact of cross-modal distillation.
However, in the pipeline of cross-modal distillation, these methods only use one modality to mimic the other which may cause \emph{imbalanced modality adaptability} in each domain.
We provide an intuitive comparison in Fig.~\ref{fig:motivation} to illustrate this limitation, 3D prediction can segment $\tt{sidewalk}$ well, whereas point-wise 2D prediction erroneously segments to $\tt{vegetation}$. In this condition, allowing 3D representations to mimic 2D representations potentially brings negative mimicking.

Motivated by the above observation, we consider how to set up a cross-modal positive distillation that benefits cross-modal learning.
It has been proven that the cross-modal fusion representation~\cite{2DPASS, UniSeg} combines the exclusive advantages of 2D images and 3D point clouds.
Images provide dense, textured, and colored information, which enriches 3D representations. On the other hand, point clouds can provide spatial-geometric structure, which supplements the spatial perception of 2D representations.
Therefore, leveraging the complementary advantage of cross-modal fusion representation to mitigate imbalanced modality adaptability in cross-modal UDA becomes a viable blueprint.
To achieve this, two challenges need to be addressed: 1) how to conduct cross-modal fusion of two heterogeneous modalities for 3D semantic segmentation and 2) how to make domain-modality alignment by exploiting cross-modal fusion representation.

As for the first challenge, considering that 2D images and 3D point clouds are heterogeneous, our objective is to achieve feature-level fusion while discarding data-level and decision-level fusion. In feature-level fusion, we intend to enforce maximum correlation and complementarity between the two heterogeneous modalities.
As for the second challenge, we consider employing cross-modal positive distillation for distribution alignment to alleviate the distribution discrepancies between different modalities and domains.
Not only in the uni-modality space, such as image and point cloud representation of intra-domain but also in the latent space formed by cross-modal fusion representation of cross-domain.

Accordingly, in this work, we propose a novel fusion-then-distillation (\textbf{FtD++}) method to explore cross-modal positive distillation of the source and target domains for 3D semantic segmentation (See Fig.~\ref{fig:motivation}(d)).
Specifically, we first implement feature-level cross-modal fusion via a model-agnostic feature fusion module (MFFM). Considering large-scale images and point clouds, features are usually refined only by calculating the self-affinities within a sample and ignoring the pair-wise relationship.
We insert several memorized modality attention blocks in MFFM, which not only can mine external affinities globally in the whole domain but also reduce computational complexity.
Then, to alleviate cross-modality and cross-domain distribution discrepancy, we design a cross-modal positive distillation (xP-Distill) learning mechanism, including modality-preserving distillation and domain-preserving distillation.
The former relieves the class probability distribution discrepancy of cross-modality by leveraging the knowledge distillation to make the output of 3D consistent with Fusion.
The latter relieves the class probability distribution discrepancy of cross-domain by leveraging ensemble knowledge distillation from the source-domain to the augment-domain.
To further illustrate, we added an evaluation of our fusion-then-distillation into Fig.~\ref{fig:per_class_iou}. Except for $\tt{sidewalk}$ where IoU declined after 60k iterations, the uptrend of per-class IoU demonstrates that our fusion-then-distillation is an effective form of positive distillation.
Additionally, we design cross-modal debiased pseudo-labeling (xDPL) during the re-training phase to mine useful supervisory information from the target-domain data. 
xDPL models the uncertainty of the pseudo-label via the multi-modal prediction variance, effectively exploiting the domain-specific information offered by pseudo-labels.

To summarize, the following are the main contributions:
\begin{itemize}
    \item [$\bullet$] We present FtD++, a novel domain adaptive 3D semantic segmentation method. It utilizes cross-modal positive distillation to explore the complementarity of cross-modal fusion representation in bridging the cross-modality and cross-domain gaps.
    \item [$\bullet$] A plug-and-play MFFM is designed for generating fusion representation, laying the groundwork for cross-modal learning. xP-Distill is proposed for cross-modal and cross-domain alignments based on the fusion representation. xDPL is intended to mine useful supervisory information from the target-domain data.
    \item [$\bullet$] Extensive experiments demonstrate that our method outperforms state-of-the-art competitors on Day$\to$Night, USA$\to$Sing., vKITTI$\to$sKITTI, and A2D2$\to$sKITTI.
\end{itemize}

Compared with the previous conference version~\cite{BFtD-xMUDA}, we mainly expand the following contents for the present work:
\begin{itemize}
    \item [$\bullet$] We extend our method to tackle semi-supervised domain adaptation (SSDA) setup. The current design allows users to input a few annotated target-domain data, which significantly improves the segmentation performance of the model on the target domain.
    \item [$\bullet$] We analyze the adaptability of cross-modal fusion representation output from MFFM and propose a cross-modal positive distillation. After re-integrating the fused features with 2D features, the hybrid prediction can undergo more effective knowledge distillation with 3D prediction, both within and across domains.
    \item [$\bullet$] We perform in-depth analysis with ablation studies on each component and provide parameter sensitivity analysis and more visualizations to highlight its strength.
\end{itemize}

\section{Related Work}

\subsection{Uni-modal 3D Semantic Segmentation}
The mainstream methods to process 3D semantic segmentation are divided into four types: Point-based method, Projection-based method, Voxel-based method, and Multi-representation method.
In the point-based method, PointNet series~\cite{PointNet, PointNet++} was a pioneering work that directly used multi-layer perceptrons (MLPs) to learn the features from the unordered sequences of point clouds.
Later on, based on PointNet, convolution operations were implemented on the point-wise features output by MLPs~\cite{PointConv, KPConv, RandLA-Net, JSNet++} that performed well on the synthetic point cloud~\cite{S3DIS, ScanNet}.
The projection-based method projected a point cloud to an image space in the bird's-eye-view or range-view, achieving efficient 3D segmentation with 2D CNNs, such as SqueezeSeg series~\cite{Squeezeseg, Squeezesegv2, Squeezesegv3} and others~\cite{RangeNet++, SalsaNext, PolarNet}.
Recently, the voxel-based method~\cite{sparseconvnet, MinkowskiNet, Cylinder3D} has been widely used for large-scale outdoor datasets, as it adopted sparse voxels to balance between efficiency and effectiveness.
Multi-representation method~\cite{SPVCNN, RPVNet, (AF)2-S3Net, PVD, YeWXCC22} used an ensemble of point-based or projection-based representation to potentially facilitate voxel representation.
However, due to the sparsity of LiDAR sensors, these methods exhibit inferior performance in segmenting distant objects.

\subsection{Cross-modal 3D Semantic Segmentation}
Image and point cloud are two heterogeneous modalities popularly used in 3D semantic segmentation for autonomous driving. Many studies have leveraged the multi-modal fusion of pixels and points to improve segmentation performance.
Early works~\cite{FuseSeg,el2019rgb} warped suboptimal RGB values or separately learned CNN features into point cloud range images as additional inputs. PMF~\cite{PMF} and Chang \textit{et al.}~\cite{ChangPSG23} projected the point cloud onto the image plane for fusion with image data through perspective projection and 2D3DNet~\cite{2D3DNet} trained a 3D model from pseudo-labels derived from 2D semantic segmentation by using multi-view fusion. 
Besides, 2DPASS~\cite{2DPASS} performed knowledge transfer from multi-modal representation to 3D features, while ProtoTransfer~\cite{ProtoTransfer} performed knowledge transfer from 2D prototype bank to 3D features directly without being restricted by matching requirements.
UniSeg~\cite{UniSeg} proposed a unified multi-modal fusion network, leveraging the information of RGB images and three views of the point cloud for more accurate and robust perception.
Although these methods have achieved remarkable performance on the condition that the distribution of training and testing samples are consistent, they may degrade significantly when the testing distribution is quite different from the training.

\begin{figure*}[t]
    \centering
    \includegraphics[width=1.0\linewidth]{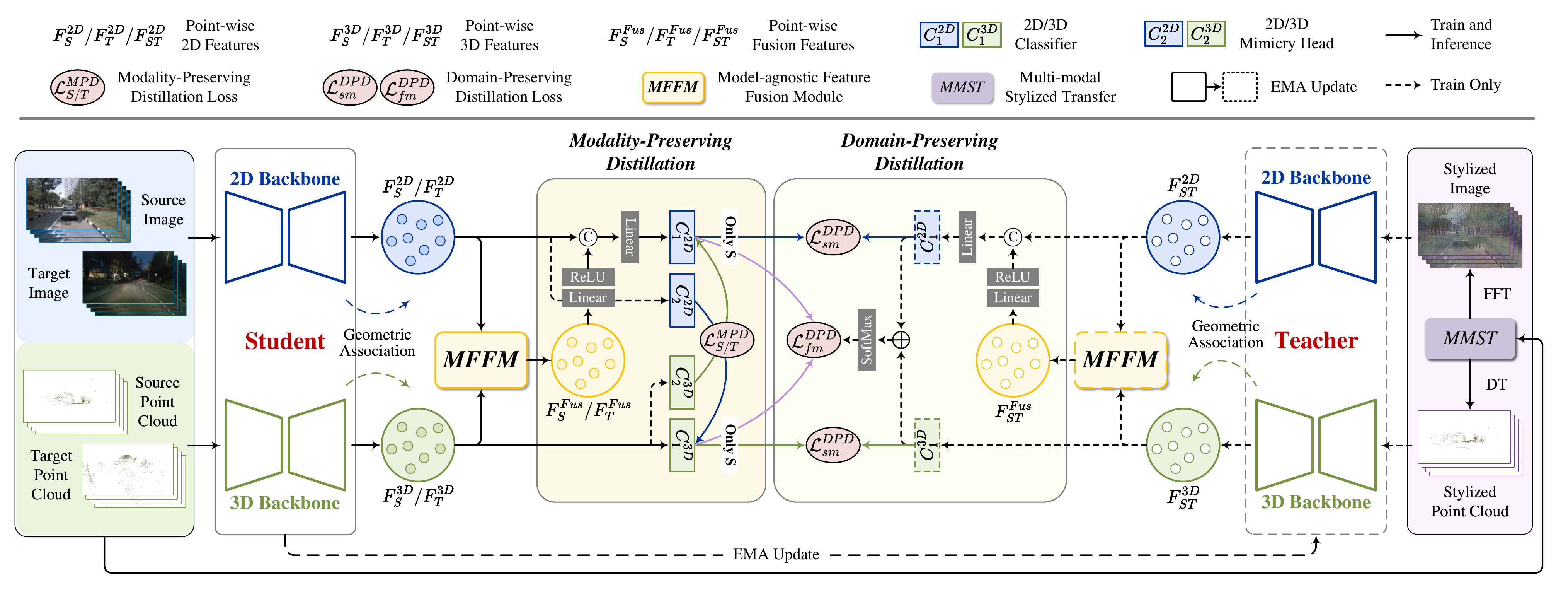}
    \caption{Overview framework of FtD++, which contains two learning mechanisms: Modality-Preserving Distillation (MPD) and Domain-Preserving Distillation (DPD). Notably, \textit{Student} and \textit{Teacher} share the same 2D and 3D backbones, but the parameters of \textit{Teacher} are updated to the counterparts of \textit{Student} via Exponential Moving Average (EMA) after each training iteration. Here, ``Only S'' means only applying source prediction for DPD.}
    \label{fig:framework}
\end{figure*}

\subsection{Uni-modal UDA for 3D Semantic Segmentation}
Uni-modal UDA seeks to narrow the domain gap between the source and target 3D domain data. Early works~\cite{ePointDA, PCT} exploited the generative adversarial network to mitigate domain shifts caused by appearance and sparsity differences.
Later on, Yuan \textit{et al.}~\cite{YuanCZSLYW23, YuanWCSLYW23} proposed the adversarial network based on category-level and prototype-level alignments to address the mismatch of sampling patterns.
CosMix~\cite{CosMix} and ConDA~\cite{KongQL23} constructed an intermediate domain by utilizing joint supervision signals from both the source and target domains for self-training.
Recently, 3D surface representation has also been considered as a solution. Complete\&Label~\cite{yi2021complete} transformed the domain adaptive task into the 3D surface completion task. SALUDA~\cite{MicheleBPVMC24} learned an implicit underlying surface representation simultaneously on source and target data.
However, constrained by the requirement for a one-to-one correspondence between pixels and points, these methods struggle to adapt to cross-modal UDA readily.

\subsection{Cross-modal UDA for 3D Semantic Segmentation}
Benefiting from the exclusive advantages of images and point clouds, cross-modal UDA becomes a research hotspot. xMUDA~\cite{xMUDA} provided a cross-modal learning method for 3D semantic segmentation in UDA.
To facilitate learning domain-robust dependencies, several methods extended 2D techniques to learn the 3D domain-invariant representations, such as adversarial learning~\cite{AUDA, DsCML, SSE-xMUDA}, style transfer~\cite{Dual-Cross}, pseudo-labeling~\cite{MM-TTA} and contrastive learning~\cite{CMCL}.
Besides, SSE~\cite{SSE-xMUDA} further presented a self-supervised learning mechanism from plane-to-spatial and discrete-to-textured representations.
MM2D3D~\cite{MM2D3D} injected depth cues into the 2D branch and color cues into the 3D branch while preserving the complementarity of predictions.
MoPA~\cite{MoPA} borrowed the knowledge priors learned from segment anything model~\cite{SAM} to produce more accurate pseudo-labels for unlabeled target domains.
Differently, we focus on utilizing the complementary advantage of cross-modality fusion to alleviate the imbalanced modality adaptability in cross-modal learning and benefit 3D semantic segmentation in UDA.

\section{Proposed Method}

\subsection{Preliminary}

\noindent \textbf{Problem Definition.} For the UDA task, we define a source domain $\mathcal{S} = \left \{ (X_{S,i}^{2D},X_{S,i}^{3D},Y_{S,i}^{3D}) \right \}_{i=1}^{n_s}$ with $n_s$ unlabeled images and labeled point clouds, and a target domain $\mathcal{T} = \left \{ (X_{T,i}^{2D},X_{T,i}^{3D}) \right \}_{i=1}^{n_t}$ with $n_t$ unlabeled images and point clouds.
Only the source point clouds have annotation $Y_{S,i}^{3D}$ belonging to $C$ classes for each 3D point.
For the SSDA task, $\mathcal{T}$ is divided into a small number of labeled target set $\mathcal{T}_{l} = \left \{ (X_{T_{l},i}^{2D},X_{T_{l},i}^{3D},Y_{T_{l},i}^{3D}) \right \}_{i=1}^{n_{tl}}$ and a large number of unlabeled target set $\mathcal{T}_{u} = \left \{ (X_{T_{u},i}^{2D},X_{T_{u},i}^{3D}) \right \}_{i=1}^{n_{tu}}$, where $n_{tl} + n_{tu} = n_{t}$.
$X^{3D}$ only contains points located at the camera reference frame, assuming that the calibration of LiDAR-Camera is available for both domains.
Note that 3D points are projected onto the image via the relative projection matrix to obtain cross-modal correspondences (\textit{i.e.,} geometric association).
Both UDA and SSDA aim to adapt 2D and 3D models trained on $\mathcal{S}$ to $\mathcal{T}$, learning a function $f:\mathcal{S} \cup \mathcal{T} \mapsto Y_{T}^{3D}$ that could predict target 3D labels.

\noindent \textbf{Revisiting xMUDA.} Generally, in xMUDA~\cite{JaritzVCWP23}, $X_{S}^{2D},X_{T}^{2D}$ and $X_{S}^{3D},X_{T}^{3D}$ are first fed into parallel 2D and 3D backbone respectively, which output image and voxel features. Accordingly, in the source and target domains, we can obtain point-wise 2D features $F_{S}^{2D},F_{T}^{2D} \in \mathbb{R}^{N \times D_{1}}$ and point-wise 3D features $F_{S}^{3D},F_{T}^{3D} \in \mathbb{R}^{N \times D_{2}}$. $N$ is the number of points projected onto the image plane, $D_{1}$ and $D_{2}$ denote the length of the 2D and 3D backbone output features. After that, classifiers $C_{1}^{2D}$ and $C_{1}^{3D}$ are used for prediction, and mimicry heads $C_{2}^{2D}$ and $C_{2}^{3D}$ are employed to estimate the other modality output via cross-modal learning. The mimicry head has the same structure as the main classifier, but the parameters are updated differently.

\subsection{Overview}
The overall framework of FtD++ is illustrated in Fig.~\ref{fig:framework}.
We first introduce a model-agnostic feature fusion module, transferring exclusive modality information from dense images and sparse point clouds to the cross-modal fusion representations.
Then we shed light on the main component of our framework, \textit{i.e.,} modality-preserving distillation for cross-modal learning and domain-preserving distillation for cross-domain learning, both of which alleviate the distribution discrepancies between different modalities and domains.
We finally introduce cross-modal debiased pseudo-labeling for the self-training scheme, using an additional segmentation loss on the target-domain training set.

\subsection{Cross-modal Feature Fusion}
How to effectively leverage cross-modal feature fusion to maximize the correlation and complementarity of 2D and 3D features remains an open challenge. To this end, a model-agnostic feature fusion module (MFFM) is introduced, which generates cross-modal fusion representation through the attention mechanism. Hereby, we take the source-domain $\mathcal{S}$ inputs as an example. Given features $F_{S}^{2D}$ and $F_{S}^{3D}$, if global self-attention~\cite{VaswaniSPUJGKP17} is adopted to $N \times D$ vectors, focusing solely on the affinity between different positions within a single sample has a high computational cost, with complexity of $O(DN^{2})$. Thus, directly applying self-attention to large-scale point-wise features is infeasible.

\begin{figure}[t]
    \centering
    \includegraphics[width=1.0\linewidth]{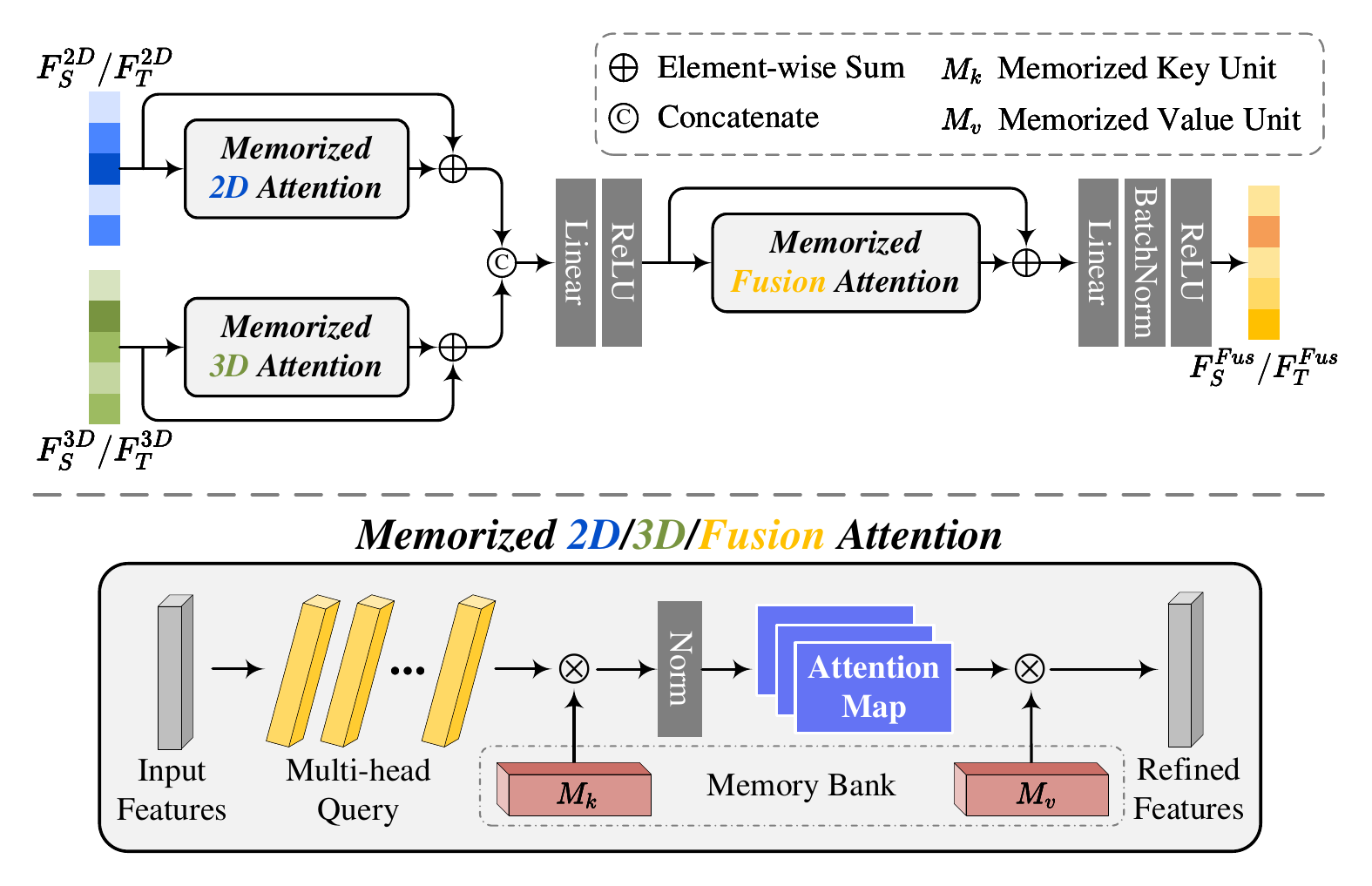}
    \caption{The diagram of MFFM with memorized modality attention modules.}
    \label{fig:fusion}
\end{figure}

Concretely, we meticulously design a memorized modality attention module that mines external affinities of multi-modality from whole domain samples. It compensates for ignoring relationships between samples across different batches in global self-attention.
Unlike global self-attention whose key and value matrices are generated from a linear projection of input, inspired by~\cite{guo2022beyond}, our module uses the learnable key unit $M_{k}^{(\cdot)}$ and value unit $M_{v}^{(\cdot)}$ and then save them in the memory bank.
If multiplying $M_{k}^{(\cdot)}$, it can learn the potential affinity between $N$ query elements and $K$ memorized key elements. If multiplying $M_{v}^{(\cdot)}$, it updates the input features from $M_{v}^{(\cdot)}$ by the similarities in attention map.
This mechanism not only mines external affinities from whole domain features but also reduces computational complexity from $O(DN^{2})$ to $O(DNK)$ due to the small neighbors but the wider scope of attention map $A^{(\cdot)} \in \mathbb{R}^{N \times K}$. As shown in Fig.~\ref{fig:fusion}, 2D and 3D attention maps are defined as:
\begin{equation}\label{eq:A_2D}
    A^{2D} = Norm(F_{S}^{2D} (M_{k}^{2D})^{\top}),
\end{equation}
\begin{equation}\label{eq:A_3D}
    A^{3D} = Norm(F_{S}^{3D} (M_{k}^{3D})^{\top}),
\end{equation}
where $A^{2D},A^{3D} \in \mathbb{R}^{N \times K}$ are the attention matrices, $M_{k}^{2D} \in \mathbb{R}^{K \times D_{1}}$ and $M_{k}^{3D} \in \mathbb{R}^{K \times D_{2}}$ are 2D and 3D memorized key units, $K$ is the length of key, $Norm(\cdot)$ is the normalization operation.
To increase the capability of the network, we calculate the pair-wise affinity between the attention matrix and value unit, which is defined as:
%
\begin{equation}\label{eq:F_2D}
    \tilde{F}_{S}^{2D} = A^{2D} M_{v}^{2D} + F_{S}^{2D},
\end{equation}
\begin{equation}\label{eq:F_3D}
    \tilde{F}_{S}^{3D} = A^{3D} M_{v}^{3D} + F_{S}^{3D},
\end{equation}
where $M_{v}^{2D} \in \mathbb{R}^{K \times D_{1}}$ and $M_{v}^{3D} \in \mathbb{R}^{K \times D_{2}}$ are 2D and 3D memorized value units. The refined 2D feature $\tilde{F}_{S}^{2D} \in \mathbb{R}^{N \times D_{1}}$ and 3D feature $\tilde{F}_{S}^{3D} \in \mathbb{R}^{N \times D_{2}}$ are strengthened.

Then, to enhance the flexibility in feature adjustment, MFFM concatenates $\tilde{F}_{S}^{2D}$ and $\tilde{F}_{S}^{3D}$, feeding them into a fusion adapter $\Phi_{1}(\cdot)$ to produce fusion feature $F_{S}^{Fus} \in \mathbb{R}^{N \times D_{f}}$, which is obtained by:
\begin{equation}\label{eq:F_Fusion}
    F_{S}^{Fus} = \Phi_{1}(\tilde{F}_{S}^{2D} \uplus \tilde{F}_{S}^{3D}),
\end{equation}
where $\uplus$ concatenates two inputs along the feature channel.

External attention~\cite{guo2022beyond} has linear complexity and implicitly considers the correlation of all samples. However, it does not involve the correlations between 2D and 3D features.
Therefore, we aim to learn exclusive and discriminative features from each modality, capturing the most informative parts and excluding interference from other samples to the fusion modality.
Specifically, after calculating $F_{S}^{Fus}$, the fusion attention module calculates external attention between the fusion representation and fusion memory units separately. Ultimately, the output refined fusion feature $\tilde{F}_{S}^{Fus}$ is obtained by:
\begin{equation}\label{eq:A_fus}
    A^{Fus} = Norm(F_{S}^{Fus} (M_{k}^{Fus})^{\top}),
\end{equation}
\begin{equation}\label{eq:F_fus}
    \tilde{F}_{S}^{Fus} = \Phi_{2}(A^{Fus} M_{v}^{Fus} + F_{S}^{Fus}),
\end{equation}
where $M_{k}^{Fus}, M_{v}^{Fus} \in \mathbb{R}^{K \times D_{f}}$ are fusion memorized key and value units, $\Phi_{2}(\cdot)$ is built with Linear, BatchNorm, and ReLU layers. In the same way, we can generate refined fusion feature $\tilde{F}_{T}^{Fus}$ of the target domain via the same MFFM.

\subsection{Cross-modal Positive Distillation}\label{sec:xP-Distill}
The mutual learning of heterogeneous modalities benefits a more robust and comprehensive perception. To this end, we establish cross-modal positive distillation, using the complementary advantage of cross-modal feature fusion to improve the domain-modality alignment, which we present below.

\noindent \textbf{Modality-Preserving Distillation.} As depicted in Fig.~\ref{fig:framework}, according to MFFM, MPD establishes a mimicking game between 2D and 3D predictions for cross-modal learning.
Unlike MPD proposed in BFtD~\cite{BFtD-xMUDA} where the outputs of 2D and 3D mimicry heads are consistent with that of the fusion classifier, MPD in FtD++ first concatenates $\tilde{F}_{S}^{Fus}$ with raw 2D feature $F_{S}^{2D}$.
Then the hybrid feature $F_{S}^{Mix}$ is fed into 2D classifier $C_{1}^{2D}$ via bottleneck structure (Linear-ReLU-Linear), toward an agreement in latent space. The final source 2D predictions are defined as follows:
\begin{equation}
    P_{S}^{2D} = C_{1}^{2D}(F_{S}^{Mix}),\quad P_{S}^{2D,m} = C_{2}^{2D}(F_{S}^{2D}),
\end{equation}
\begin{equation}
    F_{S}^{Mix} = Linear(F_{S}^{2D} \uplus ReLU(Linear(\tilde{F}_{S}^{Fus}))).
\end{equation}

Similarly, the same structure is applied to the target output, and then the final target 2D predictions are defined as:
\begin{equation}
    P_{T}^{2D} = C_{1}^{2D}(F_{T}^{Mix}),\quad P_{T}^{2D,m} = C_{2}^{2D}(F_{T}^{2D}),
\end{equation}
\begin{equation}
    F_{T}^{Mix} = Linear(F_{T}^{2D} \uplus ReLU(Linear(\tilde{F}_{T}^{Fus}))).
\end{equation}

Since images are more susceptible to the environment than point clouds, we do not consider the mixing of fusion features with 3D features, thus preserving the 3D domain predictions driven by the raw spatial correlation. Thereby, the final source and target 3D predictions are obtained by:
\begin{equation}
    P_{S}^{3D} = C_{1}^{3D}(F_{S}^{3D}),\quad P_{S}^{3D,m} = C_{2}^{3D}(F_{S}^{3D}),
\end{equation}
\begin{equation}
    P_{T}^{3D} = C_{1}^{3D}(F_{T}^{3D}),\quad P_{T}^{3D,m} = C_{2}^{3D}(F_{T}^{3D}).
\end{equation}

After that, we conduct class probability distribution alignment between 2D and 3D predictions. Hereby, we select the Kullback-Leibler (KL) divergence $D_{KL}(\cdot||\cdot)$ for the MPD loss and define it as follows:
\begin{equation}\label{eq:MPD_S}
    \mathcal{L}_{S}^{MPD} = D_{KL}(P_{S}^{2D} || P_{S}^{3D,m}) + D_{KL}(P_{S}^{3D} || P_{S}^{2D,m}),
\end{equation}
\begin{equation}\label{eq:MPD_T}
    \mathcal{L}_{T}^{MPD} = D_{KL}(P_{T}^{2D} || P_{T}^{3D,m}) + D_{KL}(P_{T}^{3D} || P_{T}^{2D,m}).
\end{equation}

\noindent \textbf{Domain-Preserving Distillation.} As depicted in Fig. \ref{fig:framework}, DPD performs cross-domain alignment via knowledge distillation eliminating the discrepancy in class probability distribution between the source and target domains.
Unlike DPD proposed in BFtD~\cite{BFtD-xMUDA} where MFFM is employed to generate the hybrid-modal prediction of a stylized sample directly, DPD in FtD++ uses hybrid-modal prediction by taking the mean value of 2D and 3D predictions output from \textit{Teacher}.

Concretely, we first adopt the multi-modal style transfer (MMST) module proposed by~\cite{Dual-Cross} to enable \textit{Teacher} to perceive the target style.
The stylized image $X_{ST}^{2D}$ assembles the low-frequency components of the target sample and the high-frequency components of the source sample, while the stylized point cloud $X_{ST}^{3D}$ transforms the density of the source point cloud to match that of the target point cloud. The stylized inputs take the following form:
\begin{equation}
    X_{ST}^{2D} = \mathcal{F}^{-1}(\mathcal{F}^{low}(X_{T}^{2D}) + \mathcal{F}^{high}(X_{S}^{2D})),
\end{equation}
\begin{equation}
    X_{ST}^{3D} = \sqrt[3]{\frac{\mathcal{D}(X_{T}^{3D})}{\mathcal{D}(X_{S}^{3D})}} X_{S}^{3D},
\end{equation}
where $\mathcal{F}^{low}(\cdot)$ and $\mathcal{F}^{high}(\cdot)$ indicate the low-frequency and high-frequency components of the Fast Fourier Transform $\mathcal{F}$ of an image, respectively, and $\mathcal{F}^{-1}$ is the inverse of $\mathcal{F}$. $\mathcal{D}(\cdot) = (X_{max}-X_{min})(Y_{max}-Y_{min})(Z_{max}-Z_{min}) / N$ indicates the density of one point cloud. $X/Y/Z_{min}$ means the minimum coordinate value while $X/Y/Z_{max}$ means the maximum.

The EMA updated teacher model realizes a temporal ensemble of the previous student model, increasing the robustness and temporal stability of perturbed sample prediction. In DPD, as \textit{Teacher} is updated based on \textit{Student}, it will gradually obtain the enhanced context learning capability from \textit{Student}. At this moment, via EMA updated \textit{Teacher} and MFFM, we can get predictions of source data in the target style:
\begin{equation}
    P_{ST}^{2D} = C_{1}^{2D}(F_{ST}^{Mix}),\quad P_{ST}^{3D} = C_{1}^{3D}(F_{ST}^{3D}),
\end{equation}
\begin{equation}
    F_{ST}^{Mix} = Linear(F_{ST}^{2D} \uplus ReLU(Linear(\tilde{F}_{ST}^{Fus}))),
\end{equation}
where $C_{1}^{2D}$ and $C_{1}^{3D}$ here are EMA updated from classifier in MPD of \textit{Student} branch.

After that, we directly enforce the probability distributions obtained by \textit{Student} and \textit{Teacher} to be consistent under the ``Source-to-Stylized'' alignment. In the source domain, \textit{Teacher} transfers the target-style information to \textit{Student}. In this way, we can make \textit{Student} perceive the target domain when trained with source domain data, improving their perception of the target domain data. Hence, DPD loss in the single 2D/3D modality is defined as:
\begin{equation}
    \mathcal{L}_{sm}^{DPD} = D_{KL}(P_{ST}^{2D} || P_{S}^{2D}) + D_{KL}(P_{ST}^{3D} || P_{S}^{3D}).
\end{equation}

Similar to $\mathcal{L}_{sm}^{DPD}$, we conduct ``Source 2D/3D to Stylized Fusion'' alignment in the fusion modality, which is defined as:
\begin{equation}
    \mathcal{L}_{fm}^{DPD} = D_{KL}(P_{ST}^{Fus} || P_{S}^{2D}) + D_{KL}(P_{ST}^{Fus} || P_{S}^{3D}),
\end{equation}
\begin{equation}
    P_{ST}^{Fus} = SoftMax(P_{ST}^{2D} + P_{ST}^{3D}).
\end{equation}
%


\subsection{Cross-modal Debiased Pseudo-Labeling}
Self-training~\cite{YangZ0S022, WuXZXWQ24} is a commonly used technique that typically employs pseudo-labels to learn from the unlabeled target data in UDA.
To address the issue of noisy pseudo-labels, we introduce cross-modal debiased pseudo-labeling (xDPL) to substitute the conventional pseudo supervision. xDPL models the uncertainty of the pseudo-label via the multi-modal prediction variance during the re-training phase.

\begin{figure}[t]
    \centering
    \includegraphics[width=1.0\linewidth]{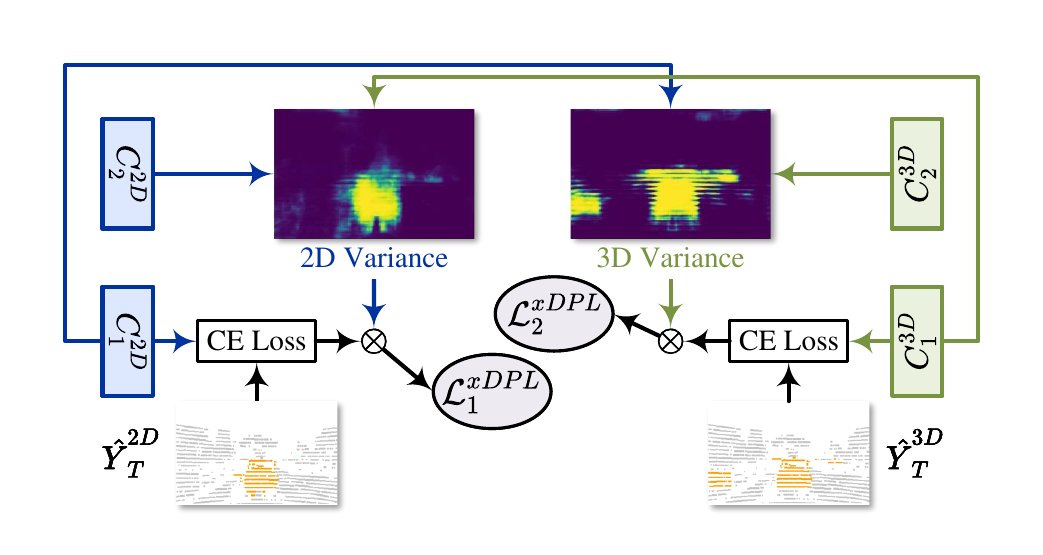}
    \caption{The diagram of xDPL for the target domain in the self-training stage. The fuzzy region denotes the high prediction variances.}
    \label{fig:xDPL}
\end{figure}
%

Inspired by~\cite{RPLL}, we utilize a variance regularization term to rectify the learning from noisy pseudo-labels, preventing the model from consistently predicting large variance.
To be specific, as shown in Fig.~\ref{fig:xDPL}, we take a 2D classifier with a 3D mimicry head ($C_{1}^{2D}, C_{2}^{3D}$) and a 3D classifier with a 2D mimicry head ($C_{1}^{3D}, C_{2}^{2D}$) to construct multi-modal prediction variance.
According to Eq. (\ref{eq:MPD_T}), KL divergence between the predictions of two heads can also be regarded as the variance. 
The approximated variance will obtain a larger value if two heads provide two different class predictions. This reflects the uncertainty of the model about the target prediction.
Given that the variance could be zero, we combine the exponential of prediction variance with the cross-entropy loss on pseudo supervision to derive the rectified unsupervised loss for both 2D and 3D terms, which can be formulated as:
\begin{equation}
    \mathcal{L}_{1}^{xDPL} = exp \big(-D_{KL}(P_{T}^{3D} || P_{T}^{2D,m}) \big) \cdot \mathcal{L}_{CE}(P_{T}^{2D},\hat{Y}_{T}^{2D}),
\end{equation}
\begin{equation}
    \mathcal{L}_{2}^{xDPL} = exp \big(-D_{KL}(P_{T}^{2D} || P_{T}^{3D,m}) \big) \cdot \mathcal{L}_{CE}(P_{T}^{3D},\hat{Y}_{T}^{3D}),
\end{equation}
where $\mathcal{L}_{CE}$ means the cross-entropy loss, $\hat{Y}_{T}^{2D}$, $\hat{Y}_{T}^{3D}$ mean the point-wise pseudo-labels of target sample generated by the model of first training stage.
In the fuzzy region, the high prediction variances enable the model to ignore its pseudo-labels, thereby reducing outliers.
In other words, xDPL provides a dynamic threshold during training, focusing the models on learning reliable pixels and points.

\subsection{Overall Loss}

The point-wise supervised segmentation loss of the source domain is formulated as follows:
\begin{equation}
    \mathcal{L}_{seg} = -\frac{1}{N \times C}\sum_{n=1}^{N} \sum_{c=1}^{C} Y_{S}^{(n,c)} \log P_{S}^{(n,c)},
\end{equation}
where $P_{S}$ is either $P_{S}^{2D}$ or $P_{S}^{3D}$. Finally, the overall loss function is defined as the sum of all the mentioned loss terms:
\begin{equation}
\begin{split}
    \mathcal{L}_{all} = & \mathcal{L}_{seg} + \lambda_{1} \mathcal{L}_{S}^{MPD} + \lambda_{2} \mathcal{L}_{T}^{MPD} + \lambda_{3} \mathcal{L}_{sm}^{DPD} + \\
    & \lambda_{4} \mathcal{L}_{fm}^{DPD} + \lambda_{5} (\mathcal{L}_{1}^{xDPL} + \mathcal{L}_{2}^{xDPL}),
\end{split}
\end{equation}
where $\{\lambda_i\}_{i=1}^5$ are weights trading off in the losses of MPD, DPD, and xDPL, respectively.

\section{Experiment}

\subsection{Datasets}
For evaluation, we use four public autonomous driving datasets, including three real scenarios: nuScenes~\cite{nuScenes}, A2D2~\cite{A2D2}, SemanticKITTI~\cite{SemanticKITTI} and one synthetic scenario: VirtualKITTI~\cite{VirtualKITTI}. For all real datasets, LiDAR and RGB cameras are synchronized and calibrated, allowing 3D-to-2D projection. For the synthetic dataset, VirtualKITTI provides depth maps so we simulate LiDAR scanning via uniform point sampling. Furthermore, we only use the front camera image and the corresponding LiDAR points.

\subsubsection{nuScenes} It contains 1,000 scenes, each of 20 seconds, corresponding to 40k annotated keyframes taken at 2Hz. The original scenes are split into 28,130 training frames and 6,019 validation frames. Each frame contains a 32-beam LiDAR point cloud with annotation and six RGB images captured by six cameras from different views of LiDAR.

\subsubsection{VirtualKITTI} It consists of 5 driving scenes which are created with the Unity game engine by real-to-virtual cloning of the scenes 1, 2, 6, 18, and 20 of the real KITTI dataset. Unlike real KITTI, VirtualKITTI does not simulate LiDAR but provides a dense depth map alongside semantic, instance, and flow ground truth. Each of the 5 scenes contains between 233 and 837 frames, \textit{i.e.,} in total 2126 for the 5 scenes. Each frame is rendered with 6 different weather or lighting variants, including clone, morning, sunset, overcast, fog, and rain.

\subsubsection{SemanticKITTI} It is a large-scale dataset based on the KITTI Odometry Benchmark captured in Germany. The original scenes are split into 19,130 training scans and 4,071 validation scans. Unlike nuScenes, SemanticKITTI only provides the front-view images and a 64-layer front LiDAR.

\subsubsection{A2D2} It consists of 20 drives corresponding to 28,637 frames. The point cloud comes from three 16-layer front LiDARs (center, left, and right), where the left and right LiDARs are inclined. By projecting 3D point clouds onto 2D images, corresponding 2D semantic labels are considered 3D point-wise labels.

\begin{table}[t]
  \centering
  \renewcommand{\arraystretch}{1.1}
  \setlength\tabcolsep{4pt}
  \caption{Size of the splits in frames for all proposed UDA and SSDA scenarios.}
  \resizebox{1.0\linewidth}{!}
  {
      \begin{tabular}{c|c|cccc}
        \toprule[1pt]
        ~ & \multirow{2}*{Adaptation Scenario} & Source & \multicolumn{2}{c}{Target} \\ \cline{3-5}
        ~ & ~ & Train & Train & Val/Test \\ \hline
        \multirow{4}*{UDA} & nuScenes: Day$\to$Night & 24745 & 2779 & 606/602  \\
        ~ & nuScenes: USA$\to$Singapore & 15695 & 9665 & 2770/2929 \\
        ~ & VirtualKITTI$\to$SemanticKITTI & 2126 & 18029 & 1101/4071 \\
        ~ & A2D2$\to$SemanticKITTI & 27695 & 18029 & 1101/4071 \\ \hline
        \multirow{3}*{SSDA} & nuScenes: USA$\to$Singapore & 15695 & \makecell{$\mathcal{T}_{l}$:2884 \\ $\mathcal{T}_{u}$:6781} & 2770/2929 \\
        ~ & A2D2$\to$SemanticKITTI & 27695 & \makecell{$\mathcal{T}_{l}$:5642 \\ $\mathcal{T}_{u}$:32738} & 1101/4071 \\
        \bottomrule[1pt]
      \end{tabular}
    }
  \label{tab:data_split}
\end{table}
\begin{table}[t]
  \centering
  \renewcommand{\arraystretch}{1.1}
  \caption{Parameters of Method Architecture.}
  \resizebox{0.9\linewidth}{!}
  {
      \begin{tabular}{c|cccc}
        \toprule[1pt]
        Arch. & 2D backbone & 3D backbone & Fusion module \\ \hline
        Model & U-Net & SparseConvNet & MFFM \\
        Params & 23.60M & 2.69M & 0.105M \\
        \bottomrule[1pt]
      \end{tabular}
    }
  \label{tab:param}
\end{table}

\subsection{Domain Adaptation Setup}
Our experimental scenarios cover three real-to-real domain adaptation challenges, including lighting changes (nuScenes: Day$\to$Night), scene layout of the countries (nuScenes: USA$\to$Sing.), and sensor setups (A2D2$\to$sKITTI).
Following the old domain adaptation setup of BFtD~\cite{BFtD-xMUDA}, we utilize the accessible 3D bounding box (``bbox'' for short) and 3D point-wise (``seg'' for short) annotations to obtain 3D semantic labels.
Particularly, the 3D bounding box annotation is based on whether the points are located in a specific 3D box and the points outside the box are labeled as $\tt{background}$. We choose 4 merged classes ($\tt{car}$, $\tt{bike}$, $\tt{pedestrian}$, $\tt{traffic\ boundary}$) with $\tt{background}$ for the first two scenarios.
Recently, cross-modal UDA works delving into segment a new split of merged classes, introduced by~\cite{JaritzVCWP23}, a journal version of xMUDA~\cite{xMUDA}. Hence, for the first two scenarios, we choose 6 merged classes ($\tt{vehicle}$, $\tt{driv.\ surf.}$, $\tt{sidewalk}$, $\tt{terrain}$, $\tt{manmade}$, and $\tt{vegetation}$), and 10 shared classes ($\tt{car}$, $\tt{truck}$, $\tt{bike}$, $\tt{person}$, $\tt{road}$, $\tt{parking}$, $\tt{sidewalk}$, $\tt{building}$, $\tt{vegetation}$, and $\tt{other\ objects}$) are selected for the last scenario.
Additionally, one synthetic-to-real domain adaptation challenge, vKITTI$\to$sKITTI, also be considered. It trains on simulated depth and RGB, adapting to real LiDAR and camera. We choose 6 merged classes between two datasets ($\tt{car}$, $\tt{truck}$, $\tt{road}$, $\tt{building}$, $\tt{vegetation}$, and $\tt{other\ objects}$). The source and target classes of all scenarios are coincident. Tab.~\ref{tab:data_split} shows the number of training, validation, and testing splits of UDA and SSDA tasks. Note that, for the UDA split of sKITTI, $\left\{ 00,01,02,03,04,05,06,09,10 \right\}$ is set as the training set, $\left\{ 07 \right\}$ as the validation set, and $\left\{ 08 \right\}$ as the testing set. For the SSDA split of sKITTI, $\left\{ 00,01 \right\}$ is set as the labeled training set, and $\left\{ 02 \cdots 06, 09 \cdots 21 \right\}$ as the unlabeled training set.

\begin{table*}[t]
  \centering
  \renewcommand{\arraystretch}{1.1}
  \caption{Quantitative results (mIoU, \%) with both uni-modal (Uni.) and cross-modal (Cross.) adaptation methods for 3D semantic segmentation on four old settings following~\cite{BFtD-xMUDA}. ``$_{PL}$'' denotes the retraining model with offline target pseudo labels. DsCML-ALL presents DsCML training with inter-domain cross-modal learning and fine-tuning with the PL. The best and second best results are marked in \textbf{bold} and \underline{underline}.}
  \resizebox{0.93\linewidth}{!}
  {
      \begin{tabular}{c|l|ccc|ccc|ccc|ccc}
        \toprule[1pt]
        \multirow{2}*{Info} & \multirow{2}*{Method} & \multicolumn{3}{c|}{Day$\to$Night (bbox)} & \multicolumn{3}{c|}{Day$\to$Night (seg)} & \multicolumn{3}{c|}{USA$\to$Sing. (bbox)} & \multicolumn{3}{c}{USA$\to$Sing. (seg)} \\ \cline{3-14}
         & & 2D & 3D & xM & 2D & 3D & xM & 2D & 3D & xM & 2D & 3D & xM \\ \hline
         & Source-only & 41.8 & 41.4 & 47.6 & 41.8 & 43.8 & 48.0 & 53.2 & 46.8 & 61.2 & 53.3 & 48.0 & 61.6 \\ \hline
        \multirow{6}*{Uni.} & MinEnt \cite{MinEnt} & 44.9 & 43.5 & 51.3 & 44.9 & 44.3 & 51.8 & 53.4 & 47.0 & 59.7 & 53.6 & 48.6 & 61.9 \\
         & FCNs in the Wild \cite{FCNs} & 42.6 & 42.3 & 47.9 & 42.6 & 43.9 & 48.7 & 53.7 & 46.8 & 61.0 & 54.0 & 49.2 & 62.4 \\
         & CyCADA \cite{CyCADA} & 45.7 & 45.2 & 49.7 & 45.5 & 47.8 & 49.6 & 54.9 & 48.7 & 61.4 & 54.9 & 51.3 & 62.6 \\
         & AdaptSegNet \cite{AdaptSegNet} & 45.3 & 44.6 & 49.6 & 45.5 & 45.3 & 49.3 & 56.3 & 47.7 & 61.8 & 56.5 & 49.0 & 62.0 \\
         & CLAN \cite{CLAN} & 45.6 & 43.7 & 49.2 & 45.6 & 45.1 & 50.1 & 57.8 & 51.2 & 62.5 & 57.7 & 52.1 & 63.1 \\
         & PL \cite{LiYV19} & 43.7 & 45.1 & 48.6 & 43.9 & 47.6 & 50.9 & 55.5 & 51.8 & 61.5 & 55.4 & 52.7 & 62.8 \\ \hline
        \multirow{18}*{Cross.} & xMUDA \cite{xMUDA} & 46.2 & 44.2 & 50.0 & 47.3 & 46.0 & 50.6 & 59.3 & 52.0 & 62.7 & 61.7 & 52.6 & 63.3 \\
         & AUDA \cite{AUDA} & 49.0 & 47.6 & 54.2 & - & - & - & 59.8 & 52.0 & 63.1 & - & - & - \\
         & DsCML \cite{DsCML} & 48.0 & 45.7 & 51.0 & 49.8 & 47.2 & 51.7 & 61.3 & 53.3 & 63.6 & 63.3 & 54.0 & 64.2 \\
         & Dual-Cross \cite{Dual-Cross} & 52.3 & 46.2 & 57.5 & 51.7 & 46.6 & 55.0 & 59.4 & 52.2 & 63.3 & 62.0 & 53.1 & 63.8 \\
         & SSE \cite{SSE-xMUDA} & 52.2 & 46.3 & 56.5 & 51.3 & 47.8 & 54.6 & 61.3 & 52.6 & \underline{65.1} & 64.0 & 53.4 & 64.6 \\
         & CMCL \cite{CMCL} & 49.0 & 46.6 & 51.6 & - & - & - & 62.0 & 54.2 & 64.5 & - & - & - \\ \cline{2-14}
         & \textbf{BFtD} \cite{BFtD-xMUDA} & 52.1 & 44.9 & \underline{59.2} & 51.2 & 46.3 & \textbf{60.9} & 59.0 & 51.8 & 64.0 & 61.4 & 55.2 & \underline{66.8} \\
         & \textbf{FtD++} & 59.2 & 46.3 & \textbf{59.4} & 58.2 & 48.0 & \underline{60.0} & \textbf{66.3} & 50.8 & 64.6 & \textbf{68.4} & 54.3 & 68.2 \\ \cline{2-14}
         & xMUDA$_{PL}$ \cite{xMUDA} & 47.1 & 46.7 & 50.8 & 48.4 & 47.5 & 51.2 & 61.1 & 54.1 & 63.2 & 63.0 & 54.3 & 64.2 \\
         & AUDA$_{PL}$ \cite{AUDA} & 48.7 & 46.2 & 55.7 & - & - & - & 59.7 & 51.7 & 63.0 & - & - & - \\
         & DsCML-ALL \cite{DsCML} & 50.1 & 48.7 & 53.0 & 51.4 & 49.8 & 53.8 & 63.9 & 56.3 & 65.1 & 65.6 & 57.5 & 66.9 \\
         & Dual-Cross$_{PL}$ \cite{Dual-Cross} & 53.6 & 46.8 & 58.2 & 52.1 & 48.2 & 56.2 & 61.5 & 54.7 & 63.8 & 64.1 & 55.7 & 65.6 \\
         & SSE$_{PL}$ \cite{SSE-xMUDA} & 52.6 & 47.0 & 56.7 & 52.2 & 48.5 & 55.9 & 64.2 & 54.6 & \underline{67.2} & 65.8 & 56.4 & 67.9 \\
         & CMCL$_{PL}$ \cite{CMCL} & 50.6 & 49.9 & 54.5 & - & - & - & 63.3 & 57.1 & 66.7 & - & - & - \\
         & MoPA$_{PL}$ \cite{MoPA} & 51.4 & 47.5 & 53.4 & - & - & - & 60.6 & 57.0 & 63.5 & - & - & - \\
         & MoPA$_{PL\times2}$ \cite{MoPA} & 52.2 & 48.8 & 54.2 & - & - & - & 61.7 & 57.2 & 64.1 & - & - & - \\ \cline{2-14}
         & \textbf{BFtD$_{PL}$} \cite{BFtD-xMUDA} & 51.8 & 48.7 & \underline{60.1} & 51.5 & 49.4 & \underline{61.4} & 60.1 & 54.0 & 65.3 & 63.5 & 57.0 & \underline{68.0} \\
         & \textbf{FtD++$_{PL}$} & 59.1 & 48.4 & \textbf{60.2} & \textbf{63.1} & 48.5 & 61.4 & \textbf{68.6} & 52.2 & 67.4 & \textbf{70.6} & 55.6 & 69.5 \\
        \bottomrule[1pt]
      \end{tabular}
    }
  \label{tab:result_1}
\end{table*}

\subsection{Implementation Details}

\subsubsection{Network}
For the 2D backbone, we use a modified version of U-Net \cite{UNet}, which consists of ResNet-34 \cite{ResNet} pre-trained on ImageNet \cite{ImageNet} as the encoder and transposed convolutions and skip connections as the decoder.
For the 3D backbone, we use the official SparseConvNet \cite{sparseconvnet} implementation. The voxel size is set to 5cm which is small enough only to have one 3D point per voxel.
The parameter of each backbone is tabulated in Tab.~\ref{tab:param}.

\subsubsection{Training}
To address the class imbalance, we employ standard 2D/3D data augmentation and log-smoothed class weights on point-wise supervised segmentation loss. Adam optimizer with $\beta_{1}$=0.9, $\beta_{2}$=0.999 is employed. The decay rate of the moving average is set to 0.999.
The batch size is set to 8 on Day$\to$Night and USA$\to$Sing., 4 on vKITTI$\to$sKITTI and A2D2$\to$sKITTI.
The 2D and 3D networks are trained with 100k/100k/60k/200k iterations for four scenarios. We use an iteration-based learning schedule where the initial learning rate is set to 1e-4 for the 2D network and 1e-3 for the 3D network. Then, they are divided by 10 at 80k-90k/80k-90k/48k-56k/160k-180k iterations.

Moreover, for Day$\to$Night and USA$\to$Sing., $\lambda_{1}$ and $\lambda_{2}$ in MPD are set to 1.0 and 0.1, while $\lambda_{3}$ and $\lambda_{4}$ in DPD are set to 0.1 and 0.01.
For vKITTI$\to$sKITTI and A2D2$\to$sKITTI, $\lambda_{1}$ and $\lambda_{2}$ in MPD are set to 0.1 and 0.02, while $\lambda_{3}$ and $\lambda_{4}$ in DPD are set to 0.01 and 0.01.
Note that we initially train the network without using xDPL loss, \textit{i.e.,} $\lambda_{5}=0$, and $\lambda_{5}=1.0$ is an empirical value in the self-training stage.
At each iteration, we compute and accumulate gradients on the source and target batch, jointly training the 2D and 3D backbones. Our model is trained and evaluated on a GeForce RTX 3090 GPU with 24GB RAM. It takes about 2 to 4 days to train our models.

\subsubsection{Evaluation Metric}
We adopt the Intersection-over-Union (IoU) of each class and mIoU of all classes as the evaluation metric. The IoU of class $c$ is calculated via $IoU_{c}=\frac{TP_{c}}{TP_{c}+FP_{c}+FN_{c}}$, where $TP_{c}$, $FP_{c}$, and $FN_{c}$ denote the true positive, false positive, and false negative of class $c$, respectively.
In addition to evaluating 2D and 3D mIoU results, we present the ensemble result ``xM'', obtained by averaging the predicted probabilities from the 2D and 3D branches. In this work, we take the maximum value as the final result.

\begin{table*}[t]
  \centering
  \renewcommand{\arraystretch}{1.1}
  \caption{Quantitative results (mIoU, \%) with both uni-modal (Uni.) and cross-modal (Cross.) adaptation methods for 3D semantic segmentation on four new settings following~\cite{JaritzVCWP23}. ``$\dagger$'' indicates the reproduced results by using their official code.}
  \resizebox{0.93\linewidth}{!}
  {
    \begin{tabular}{c|l|ccc|ccc|ccc|ccc}
        \toprule[1pt]
        \multirow{2}*{Info} & \multirow{2}*{Method} & \multicolumn{3}{c|}{Day$\to$Night} & \multicolumn{3}{c|}{USA$\to$Sing.} & \multicolumn{3}{c|}{vKITTI$\to$sKITTI} & \multicolumn{3}{c}{A2D2$\to$sKITTI} \\ \cline{3-14}
         & & 2D & 3D & xM & 2D & 3D & xM & 2D & 3D & xM & 2D & 3D & xM \\ \hline
        ~ & Source-only & 47.8 & 68.8 & 63.3 & 58.4 & 62.8 & 68.2 & 26.8 & 42.0 & 42.2 & 34.2 & 35.9 & 40.4 \\ \hline
        \multirow{3}*{Uni.} & Deep logCORAL \cite{MorerioCM18} & 47.7 & 68.7 & 63.7 & 64.4 & 63.2 & 69.4 & 41.4 & 36.8 & 47.0 & 35.1 & 41.0 & 42.2 \\
        ~ & MinEnt \cite{MinEnt} & 47.1 & 68.8 & 63.6 & 57.6 & 61.5 & 66.0 & 39.2 & 43.3 & 47.1 & 37.8 & 39.6 & 42.6 \\ \cline{2-14}
        ~ & PL \cite{LiYV19} & 47.0 & 69.6 & 63.0 & 62.0 & 64.8 & 70.4 & 21.5 & 44.3 & 35.6 & 34.7 & 41.7 & 45.2 \\ \hline
        \multirow{14}*{Cross.} & xMUDA \cite{JaritzVCWP23} & 55.5 & 69.2 & 67.4 & 64.4 & 63.2 & 69.4 & 42.1 & 46.7 & 48.2 & 38.3 & 46.0 & 44.0 \\
        ~ & AUDA$^{\dagger}$ \cite{AUDA} & 55.6 & 69.8 & 64.8 & 64.0 & 64.0 & 69.2 & 35.8 & 37.8 & 41.3 & 43.0 & 43.6 & 46.8 \\
        ~ & DsCML$^{\dagger}$ \cite{DsCML} & 50.9 & 49.3 & 53.2 & 65.6 & 56.2 & 66.1 & 38.4 & 38.4 & 45.5 & 39.6 & 45.1 & 44.5 \\
        ~ & Dual-Cross$^{\dagger}$ \cite{Dual-Cross} & 58.5 & 69.7 & 68.0 & 64.7 & 58.1 & 66.5 & 40.7 & 35.1 & 44.2 & 44.3 & 46.1 & 48.6 \\
        ~ & SSE$^{\dagger}$ \cite{SSE-xMUDA} & 62.8 & 69.0 & \underline{68.9} & 64.9 & 63.9 & 69.2 & 45.9 & 40.0 & 49.6 & 44.5 & 46.8 & 48.4 \\ \cline{2-14}
        ~ & \textbf{BFtD}$^{\dagger}$ \cite{BFtD-xMUDA} & 57.1 & 70.4 & 68.3 & 63.7 & 62.2 & \underline{69.4} & 41.5 & 45.5 & \underline{51.5} & 40.5 & 44.4 & \underline{48.7} \\
        ~ & \textbf{FtD++} & 68.8 & 69.6 & \textbf{71.0} & 69.7 & 64.6 & \textbf{69.8} & 51.0 & 44.0 & \textbf{52.6} & 48.8 & 46.2 & \textbf{51.1} \\ \cline{2-14}
        ~ & xMUDA$_{PL}$~\cite{JaritzVCWP23} & 57.6 & 69.6 & 64.4 & 67.0 & 65.4 & 71.2 & 45.8 & 51.4 & 52.0 & 41.2 & 49.8 & 47.5 \\
        ~ & AUDA$_{PL}$$^{\dagger}$\cite{AUDA} & 54.3 & 69.6 & 61.1 & 65.9 & 65.3 & 70.6 & 35.9 & 45.5 & 45.9 & 46.8 & 48.1 & 50.6 \\
        ~ & DsCML$_{PL}$$^{\dagger}$\cite{DsCML} & 51.4 & 49.8 & 53.8 & 65.6 & 57.5 & 66.9 & 39.6 & 41.8 & 42.2 & 46.8 & 51.8 & 52.4 \\
        ~ & Dual-Cross$_{PL}$$^{\dagger}$\cite{Dual-Cross} & 59.1 & 69.0 & \underline{68.2} & 66.5 & 59.8 & 68.8 & 43.1 & 39.4 & 47.6 & 44.9 & 52.8 & 52.3 \\
        ~ & SSE$_{PL}$$^{\dagger}$\cite{SSE-xMUDA} & 59.1 & 67.0 & 66.3 & 66.9 & 64.4 & 70.6 & 47.2 & 53.5 & 55.2 & 45.9 & 51.5 & 52.5 \\ \cline{2-14}
        ~ & \textbf{BFtD$_{PL}$}$^{\dagger}$ \cite{BFtD-xMUDA} & 60.6 & 70.0 & 66.6 & 65.9 & 66.0 & \underline{71.3} & 48.6 & 55.4 & \underline{57.5} & 42.6 & 53.7 & \underline{52.7} \\
        ~ & \textbf{FtD++$_{PL}$} & 68.9 & 70.3 & \textbf{71.8} & 71.7 & 65.6 & \textbf{72.3} & 52.9 & 51.2 & \textbf{57.8} & 51.4 & 49.7 & \textbf{54.8} \\ \hline
        ~ & Target-only & 61.5 & 69.8 & 69.2 & 75.4 & 76.0 & 79.6 & 66.3 & 78.4 & 80.1 & 59.3 & 71.9 & 73.6 \\
        \bottomrule[1pt]
    \end{tabular}
  }
  \label{tab:result_2}
\end{table*}

\subsection{Main Results}
In this subsection, we divide the experimental comparisons of cross-modal UDA into two parts: \underline{Part I} is based on the old domain adaptation setup of \cite{xMUDA}, and the other \underline{Part II} is based on the new domain adaptation setup of~\cite{JaritzVCWP23}. Besides, we follow protocol~\cite{JaritzVCWP23} to evaluate the results of SSDA in \underline{Part III}, ensuring the scalability of the proposed method. 

\subsubsection{Part I}
We compare our method with six typical 2D UDA methods~\cite{MinEnt, LiYV19, FCNs, CyCADA, AdaptSegNet, CLAN}, which can be easily extended to solve cross-modal UDA. Moreover, eight cross-modal UDA methods~\cite{xMUDA, AUDA, DsCML, Dual-Cross, SSE-xMUDA, CMCL, MoPA, BFtD-xMUDA} are discussed. The comparison results for 3D semantic segmentation on the target testing data are shown in Tab.~\ref{tab:result_1}. Overall, FtD++ performs best against all competitors.
The source-only model is the lower bound, which is not domain adaptation as it is only trained on the source-domain dataset. Our method brings a significant adaptation effect on four scenarios compared to the source-only model, with gains of +11.8\%, +12.0\%, +5.1\%, and +6.8\%.
Compared with the baseline (xMUDA), FtD++ exceeds it by large margins with gains of +9.4\%, +9.4\%, +3.6\%, and +5.1\%. Compared with BFtD~\cite{BFtD-xMUDA}, it is observed that FtD++ achieves performance with +0.2\%, -0.9\%, +2.3\%, and +1.6\% gains, which implies that finding the affinity between the fusion modality and uni-modality facilitates efficient positive distillation.
Furthermore, cross-modal learning and self-training with pseudo-labels (PL) are complementary. When re-training the model with offline target PL, our method remains competitive.
It is observed that the most significant improvements with +0.1\%, +1.7\%, +3.3\%, and +2.6\% upon BFtD, which implies that re-integrating the fused features with 2D features also contributes to the self-training process by suppressing pseudo-label noise.

%
\begin{figure*}[t]
    \centering
    \includegraphics[width=0.90\linewidth]{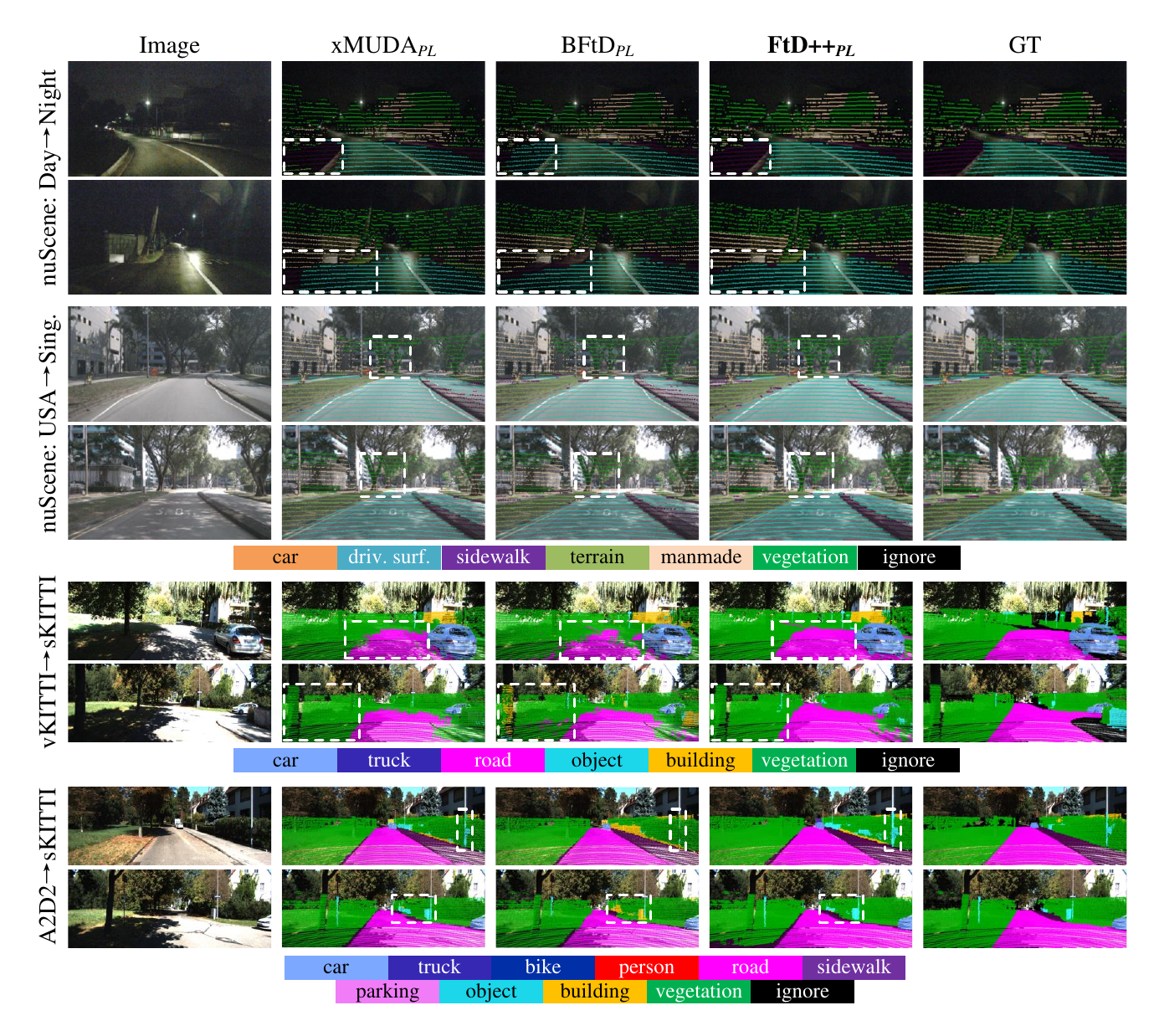}
    \caption{Qualitative results of cross-modal predictions. We present the ensembling results of ``xM'' on the target test set of four adaptive scenarios. Here, different colors denote different semantic classes of point-level prediction. The differences are highlighted using dashed rectangular boxes. Here, ``ignore'' refers to other classes that do not simultaneously exist in the ground truth (GT) of both the source and target domains.}
    \label{fig:viz}
\end{figure*}

\subsubsection{Part II}
We compare our method with three typical 2D UDA methods~\cite{MorerioCM18, MinEnt, LiYV19}, which can be easily extended to solve cross-modal UDA. Moreover, six cross-modal UDA methods~\cite{xMUDA, AUDA, DsCML, Dual-Cross, SSE-xMUDA} are discussed.
The comparison results for 3D semantic segmentation on the target testing data are shown in Tab.~\ref{tab:result_2}. Undoubtedly, FtD++ and FtD++$_{PL}$ perform best against all uni-modal and cross-modal competitors.
Similarly, it is observed that our method brings a significant adaptation effect on four scenarios compared to the source-only model, with gains of +7.7\%, +1.6\%, +10.4\%, and +10.7\%.
Compared with BFtD~\cite{BFtD-xMUDA}, the new version FtD++ achieves superior performance with +2.7\%, +0.4\%, +1.1\%, and +2.4\% mIoU gains. The results demonstrate the effect of the proposed novel combination strategy for fusion representation compared to the direct fusion prediction.
Moreover, re-training models with FtD++$_{PL}$ exceeds BFtD$_{PL}$ by large margins with gains of +5.2\%, +1.0\%, +0.3\%, and +2.1\%.
In particular, on Day$\to$Night, cross-modal positive distillation enables notable enhancing the robustness of 2D predictions, guarding against excessive distribution discrepancies that could lead to imbalanced modality adaptability.

\subsubsection{Part III}
We compare our method with three typical 2D UDA methods~\cite{MorerioCM18, MinEnt, LiYV19}, which can be easily extended to solve cross-modal SSDA. Additionally, we simply compare with xMoSSDA~\cite{JaritzVCWP23}, a semi-supervised condition of xMUDA, without introducing any semi-supervised learning strategies to ensure fairness.
The comparison results for 3D semantic segmentation on the target testing data are shown in Tab.~\ref{tab:result_3}.
Overall, FtD++ and FtD++$_{PL}$ perform relative improvement against all competitors, \textit{i.e.,} +1.3\% and +1.8\% gains on USA$\to$Sing., +4.8\% and +2.5\% gains on A2D2$\to$sKITTI compared to xMoSSDA. 
Here, ``PL'' means re-training the model with unlabeled target pseudo-labels.
Notably, FtD++ achieves performance (70.7\%) on par with xMoSSDA$_{PL}$ on A2D2$\to$sKITTI, without introducing PL, supporting the point that a small amount of supervision from the target-domain data helps the MFFM extract robust visual features.
Furthermore, FtD++$_{PL}$ is close to, or even outperforms fully supervised learning on the target-domain data alone (``Target-only'' shown in Tab.~\ref{tab:result_2}), which indirectly highlights the importance of cross-domain transfer learning (\textit{i.e.,} learn from large-scale source-domain data).

\begin{table}[t]
  \centering
  \renewcommand{\arraystretch}{1.1}
  \setlength\tabcolsep{3pt}
  \caption{Quantitative results (mIoU, \%) with domain adaptive 3D semantic segmentation under SSDA setting. We have source-domain $\mathcal{S}$, a small number of labeled target-domain $\mathcal{T}_{l}$, and a large number of unlabeled target-domain $\mathcal{T}_{u}$.}
  \resizebox{1.0\linewidth}{!}
  {
      \begin{tabular}{lc|ccc|ccc}
        \toprule[1pt]
        \multirow{2}*{Method} & \multirow{2}*{Training Set} & \multicolumn{3}{c|}{USA$\to$Sing.} & \multicolumn{3}{c}{A2D2$\to$sKITTI} \\ \cline{3-8}
        ~ & ~ & 2D & 3D & xM & 2D & 3D & xM \\ \hline
        Labeled Trg-only & $\mathcal{T}_{l}$ & 70.5 & 74.1 & 74.2 & 51.3 & 57.7 & 59.2 \\
        Src \& Labeled Trg & $\mathcal{S}+\mathcal{T}_{l}$ & 72.3 & 73.1 & 78.1 & 54.8 & 62.4 & 66.2 \\ \hline
        Deep logCORAL~\cite{MorerioCM18} & $\mathcal{S}+\mathcal{T}_{l}+\mathcal{T}_{u}$ & 71.7 & 73.1 & 78.2 & 55.1 & 62.2 & 64.7 \\
        MinEnt~\cite{MinEnt} & $\mathcal{S}+\mathcal{T}_{l}+\mathcal{T}_{u}$ & 72.6 & 73.3 & 76.6 & 56.3 & 62.5 & 65.0 \\
        xMoSSDA~\cite{JaritzVCWP23} & $\mathcal{S}+\mathcal{T}_{l}+\mathcal{T}_{u}$ & 74.3 & 74.1 & 78.5 & 56.5 & 63.4 & 65.9 \\
        \textbf{FtD++} & $\mathcal{S}+\mathcal{T}_{l}+\mathcal{T}_{u}$ & 78.4 & 73.5 & \textbf{79.8} & 69.8 & 63.5 & \textbf{70.7} \\ \hline
        PL~\cite{LiYV19} & $\mathcal{S}+\mathcal{T}_{l}+\mathcal{T}_{u}$ & 73.6 & 74.4 & 79.3 & 57.2 & 66.9 & 68.5 \\
        xMoSSDA$_{PL}$~\cite{JaritzVCWP23} & $\mathcal{S}+\mathcal{T}_{l}+\mathcal{T}_{u}$ & 75.5 & 74.8 & 78.8 & 59.1 & 68.2 & 70.7 \\
        \textbf{FtD++$_{PL}$} & $\mathcal{S}+\mathcal{T}_{l}+\mathcal{T}_{u}$ & 79.7 & 74.3 & \textbf{80.6} & 71.8 & 65.5 & \textbf{73.2} \\
        \bottomrule[1pt]
      \end{tabular}
    }
  \label{tab:result_3}
\end{table}

\noindent\textbf{Visualization.} Fig.~\ref{fig:viz} shows the qualitative segmentation results of our FtD++$_{PL}$ compared with the baseline method xMUDA$_{PL}$ and conference version BFtD$_{PL}$ with front view across all scenarios. For Day$\to$Night where xMUDA$_{PL}$ and BFtD$_{PL}$ struggle to divide the boundary between $\tt{driv.\ surf.}$ and $\tt{sidewalk}$, our FtD++$_{PL}$ can smoothly distinguish them from near to far. For USA$\to$Sing. and vKITTI$\to$sKITTI, where xMUDA$_{PL}$ and BFtD$_{PL}$ struggle to segment out the shape of $\tt{vegetation}$, our FtD++$_{PL}$ produces more accurate predictions with fewer true negatives. For the more challenging scan from A2D2$\to$sKITTI, our FtD++$_{PL}$ surpasses xMUDA$_{PL}$ and BFtD$_{PL}$ in segmenting the shape of $\tt{object}$ located far from the sensors.

\begin{table}[t]
  \centering
  \renewcommand{\arraystretch}{1.1}
  \setlength\tabcolsep{5pt}
  \caption{Ablation study on MFFM. \#x denotes the x-th model.}
  \resizebox{0.9\linewidth}{!}
  {
      \begin{tabular}{cc|ccc|ccc}
        \toprule[1pt]
        \multirow{2}*{~} & \multirow{2}*{MFFM} & \multicolumn{3}{c|}{Day$\to$Night (Seg)} & \multicolumn{3}{c}{USA$\to$Sing. (Seg)} \\ \cline{3-8}
        ~ & ~ & 2D & 3D & xM & 2D & 3D & xM \\ \hline
        \#1 & w/o att. & 56.3 & 45.7 & 57.7 & 66.4 & 54.0 & 65.6 \\ \hline
        \#2 & Single att. & 57.3 & 47.1 & 58.5 & 67.5 & 54.0 & 67.1 \\
        \#3 & Fusion att. & 57.9 & 47.4 & 58.9 & 67.7 & 54.3 & 67.2 \\
        \#4 & \#2 \& \#3 & 58.2 & 48.0 & 60.0 & 68.4 & 54.3 & 68.2 \\
        \bottomrule[1pt]
      \end{tabular}
    }
  \label{tab:ablation_MFFM}
\end{table}

\subsection{Ablation Studies}
Here, we will give a detailed analysis of the components of our method. The model training is relatively unstable, so we ran each ablation study 3 times and took the best performance.

\subsubsection{Effect of MFFM}
As shown in Tab.~\ref{tab:ablation_MFFM}, we analyze the impact of placing the memorized modality attention module at different positions in MFFM. First, we concatenate point-wise 2D and 3D features and directly feed them into adapter $\Phi_2$, without adopting any attention mechanism (\#1).
Then, we place the memorized modality attention module at different positions before features are fed into $\Phi_2$. In \#2: the modules are attached to 2D and 3D backbones. It outperforms the baseline with the gain of +0.8\% and +1.5\%. In \#3: the module is applied to the concatenated features applied in \#1. It achieves the significant gains of +1.2\% and +1.6\%.
Finally, MFFM consists of \#2 and \#3, achieving the optimal results. These experimental results show that mutual learning under cross-modal fusion representation can effectively mitigate the domain gap issue.

\begin{table}[t]
  \centering
  \renewcommand{\arraystretch}{1.1}
  \caption{Ablation study on xP-Distill. \#x denotes the x-th model.}
  \setlength\tabcolsep{4pt}
  \resizebox{1.0\linewidth}{!}
  {
      \begin{tabular}{cccc|ccc|ccc}
        \toprule[1pt]
        \multirow{2}*{~} & \multirow{2}*{MPD} & \multirow{2}*{DPD} & \multirow{2}*{MMST} & \multicolumn{3}{c|}{Day$\to$Night (Seg)} & \multicolumn{3}{c}{USA$\to$Sing. (Seg)} \\ \cline{5-10}
        ~ & ~ & ~ & ~ & 2D & 3D & xM & 2D & 3D & xM \\ \hline
        \#1 & \multicolumn{3}{c|}{xMUDA} & 47.3 & 46.0 & 50.6 & 61.7 & 52.6 & 63.3 \\ \hline
        \#2 & $\checkmark$ & ~ & ~  & 57.2 & 45.5 & 58.1 & 66.3 & 51.7 & 66.2 \\
        \#3 & $\checkmark$ & $\checkmark$ & ~ & 57.9 & 46.3 & 59.3 & 67.3 & 52.9 & 67.0 \\
        \#4 & $\checkmark$ & $\checkmark$ & $\checkmark$ & 58.2 & 48.0 & 60.0 & 68.4 & 54.3 & 68.2 \\
        \bottomrule[1pt]
      \end{tabular}
  }
  \label{tab:ablation_comp}
\end{table}

\subsubsection{Effect of xP-Distill}
As shown in Tab.~\ref{tab:ablation_comp}, xMUDA~\cite{JaritzVCWP23} is the baseline which uses naive knowledge distillation to achieve cross-modal learning (\#1).
MFFM is an indispensable core module for implementing xP-Distill (\#2, \#3, and \#4). Based on the cross-modal fusion representation of MFFM, we conduct an ablation study to analyze the two distillation mechanisms.
Firstly, in \#2, we simply add MPD to the baseline, which improves the results by +7.5\% and +3.0\%. The results indicate that transferring exclusive modality information from dense images and sparse point clouds to cross-modal fusion representations is beneficial for cross-modal learning.
Following this, in \#3, considering the effect of DPD on cross-modal learning when only raw source domain samples are fed into EMA updated \textit{Teacher} in DPD, it can still bring an extra gain of +1.2\% and +1.0\%.
Finally, in \#4, adding the MMST module for the source sample in the target style, our method achieves the best results with mIoU of 60.0\% and 68.4\%.


%
\begin{figure}[t]
    \centering
    \includegraphics[width=1.0\linewidth]{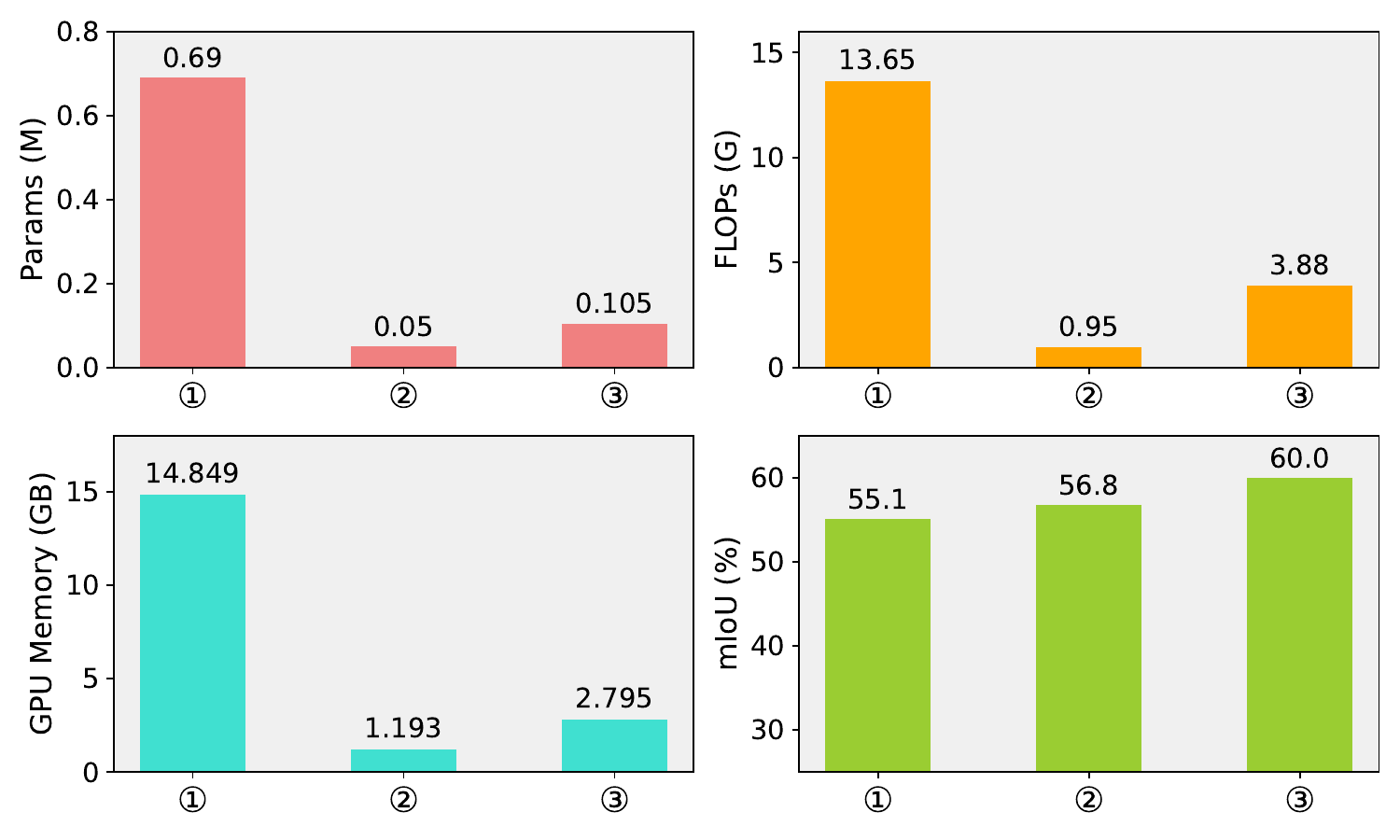}
    \caption{Computation complexity of different attention modules, including \textcircled{1}: self-attention; \textcircled{2}: channel attention; \textcircled{3}: memorized attention (Ours).}
    \label{fig:complexity}
\end{figure}

\subsubsection{Computation Complexity of Attention Module}
To evaluate the cost of the proposed fusion module, in Fig.~\ref{fig:complexity}, we show the comparison of computational complexity between self-attention~\cite{VaswaniSPUJGKP17}, channel attention~\cite{HuSS18}, and our proposed memorized modality attention. It is observed that our method with memorized attention achieves better results while keeping the floating point per second (FLOPs), model parameters (Params), and GPU memory relatively low.

%
\begin{figure}[t]
    \centering
    \includegraphics[width=0.95\linewidth]{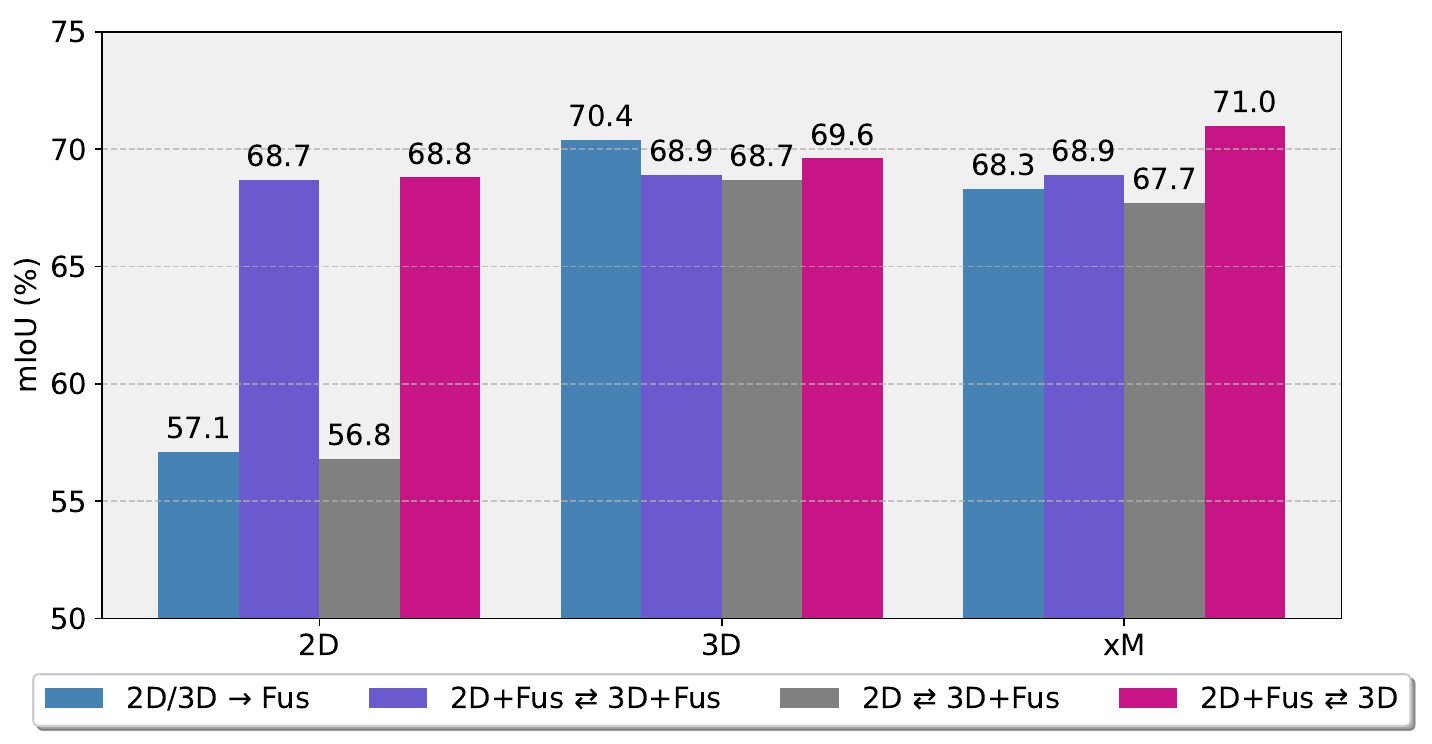}
    \caption{Analysis of fusion directivity on modal evaluation metrics, \textit{i.e.,} 2D, 3D, and xM, where ``2D+Fus$\rightleftharpoons$3D'' is the best strategy for this work.}
    \label{fig:direction}
\end{figure}

\subsubsection{Effect of Fusion Directivity}
As it can be seen from Tab.~\ref{tab:result_1} and Tab.~\ref{tab:result_2}, BFtD~\cite{BFtD-xMUDA} performs well in the old setting but shows poorer ability in the new setting with different categories.
As shown in Fig.~\ref{fig:direction}, ``2D/3D$\to$Fus'' means the bidirectional distillation in BFtD, which established a mimicking game between the 2D/3D and fusion predictions.
We found that when the 2D and 3D predictions differ significantly, their fusion effect is far inferior to the prediction of the stronger modality (\textit{i.e.,} 3D).
Hereby, an issue emerged: \textit{whether both 2D and 3D predictions need to converge towards the fusion prediction}. Therefore, we conduct the following analysis.
Firstly, we evaluate the compatibility of the fusion representation with the uni-modality representation (\textit{i.e.,} 2D+Fus$\rightleftharpoons$3D+Fus).
Before using the classifier for prediction, we concatenate the fused features with 2D and 3D features to generate hybrid features, facilitating mutual learning between the two hybrid predictions.
It is observed that the fusion feature benefits the prediction of the 2D branch but decreases the prediction performance of the 3D branch.
Subsequently, the fusion feature is concatenated with the 2D (or 3D) features, while the 3D (or 2D) modality retains its original feature representation (\textit{i.e.,} 2D$\rightleftharpoons$3D+Fus and 2D+Fus$\rightleftharpoons$3D). 
Experimental results demonstrate that cross-modal positive distillation enables notable enhancing the robustness of 2D predictions, guarding against excessive distribution discrepancies that could lead to imbalanced modality adaptability.

\begin{figure*}[t]
    \renewcommand{\figurename}{Figure}
    \centering
    \subfloat[2D style transfer on Day$\to$Night]{\includegraphics[width=0.3\linewidth]{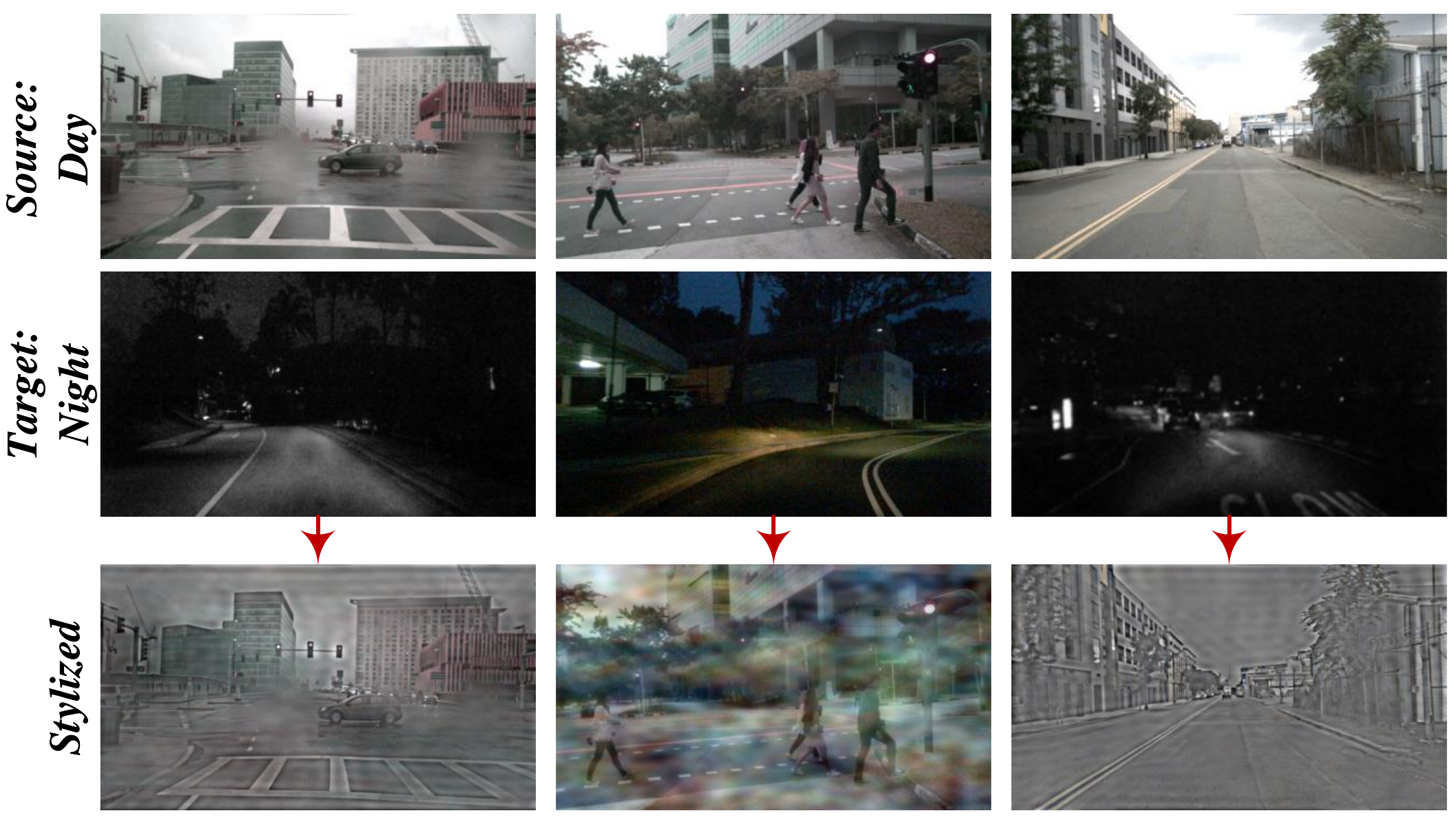}
    \label{fig:mmst_dn_2d}}
    \quad
    \subfloat[2D style transfer on USA$\to$Sing.]{\includegraphics[width=0.3\linewidth]{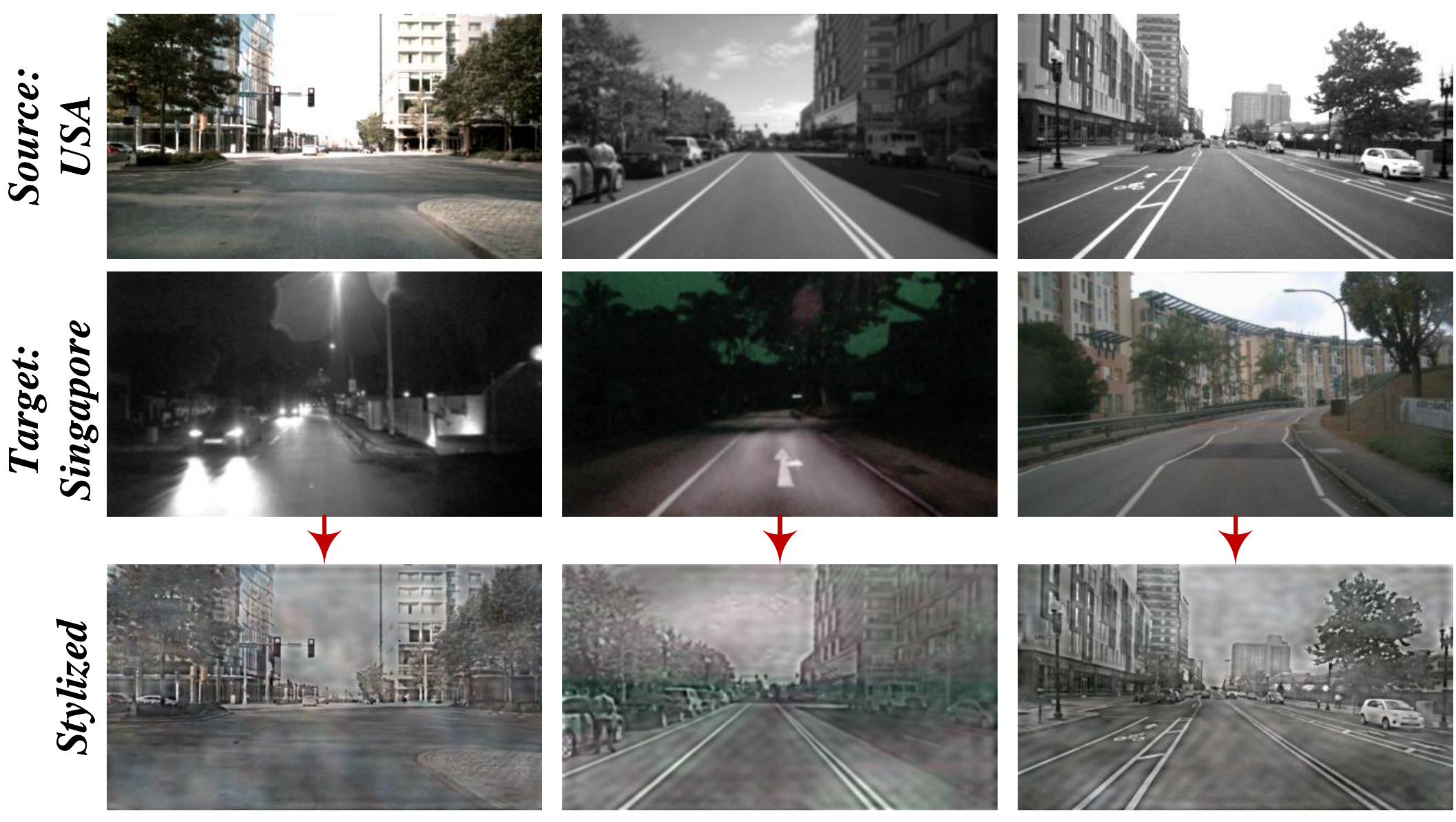}
    \label{fig:mmst_us_2d}}
    \quad
    \subfloat[2D style transfer on A2D2$\to$sKITTI]{\includegraphics[width=0.3\linewidth]{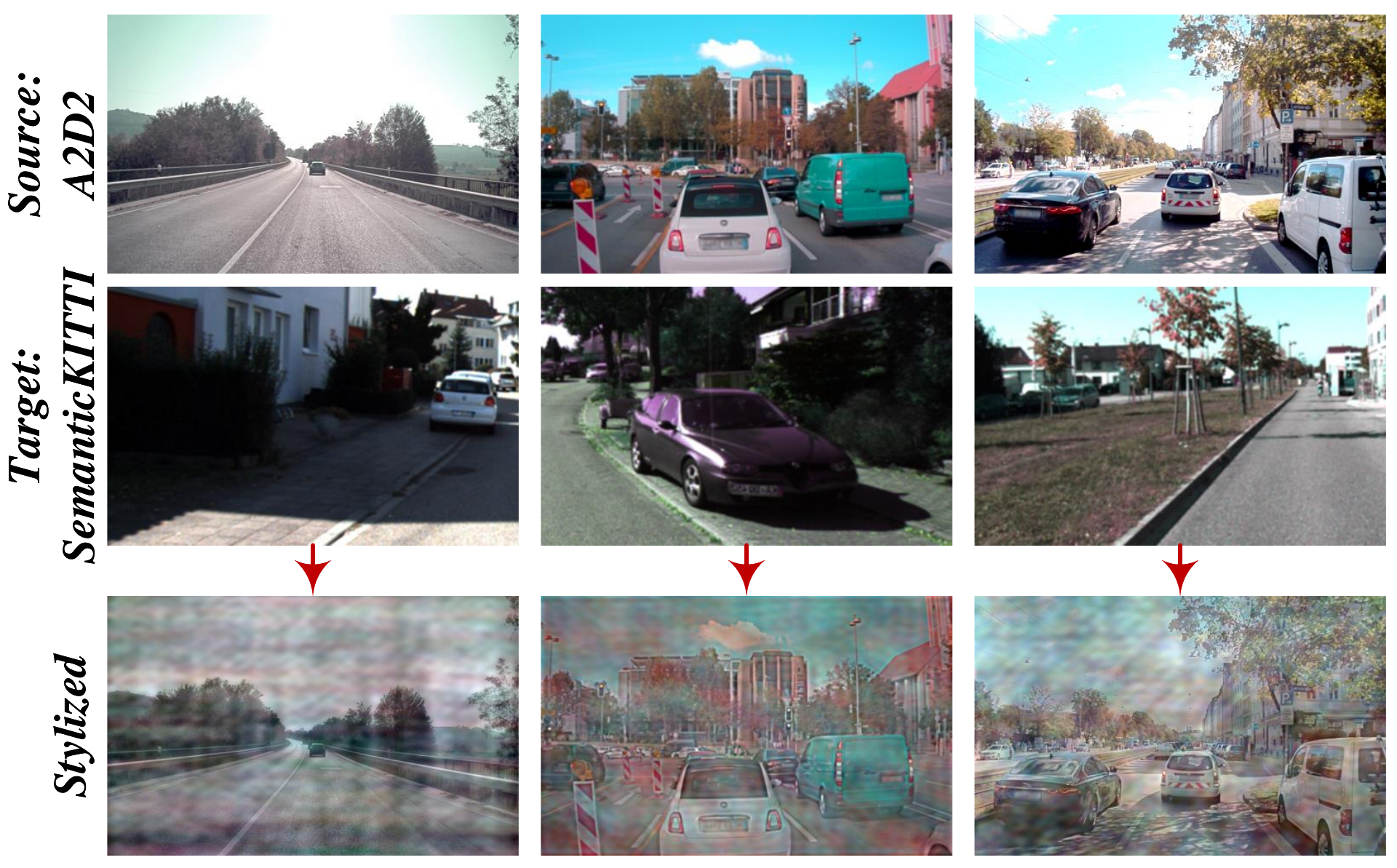}
    \label{fig:mmst_as_2d}}
    \quad
    \subfloat[3D style transfer on Day$\to$Night]{\includegraphics[width=0.3\linewidth]{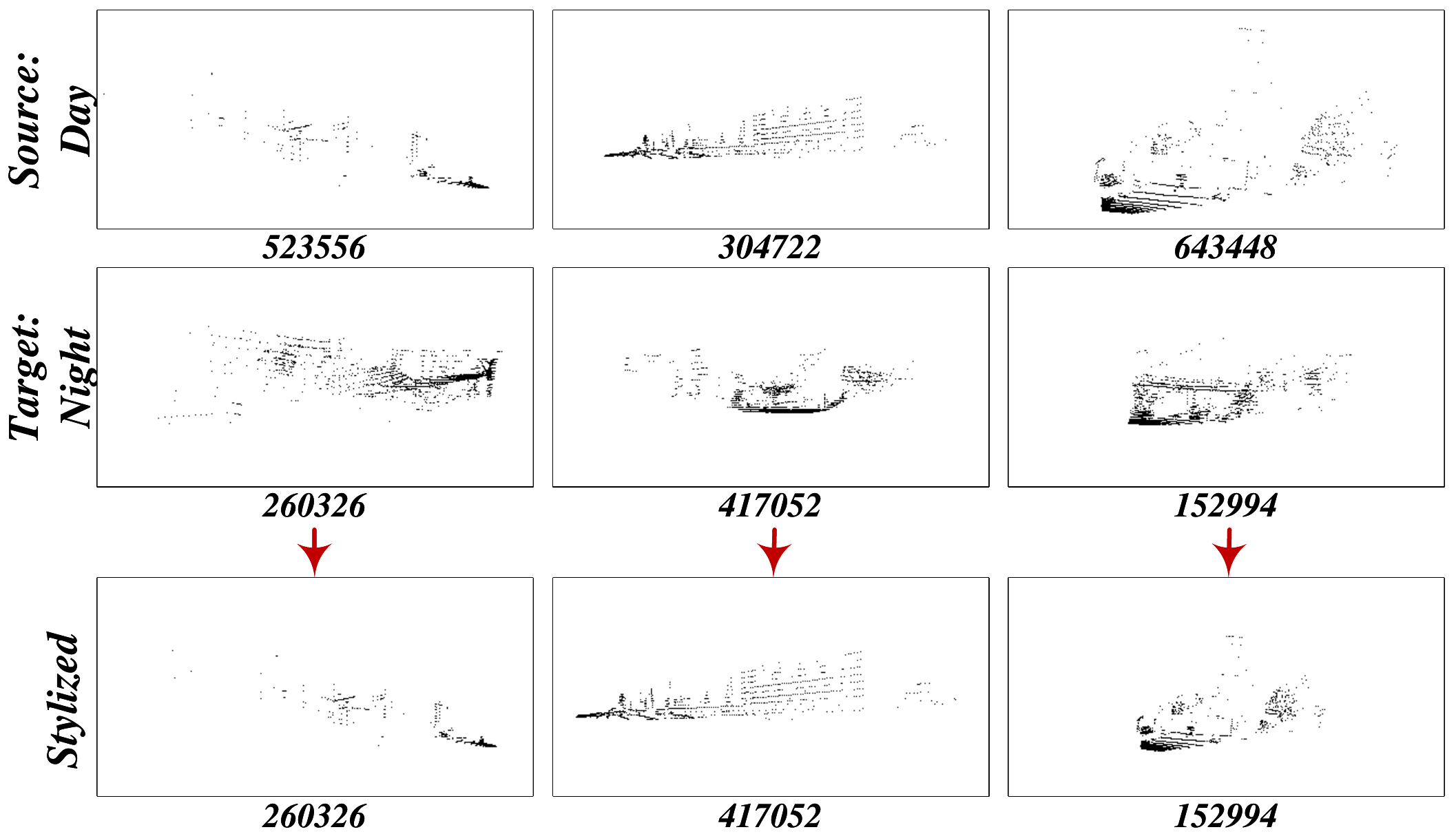}
    \label{fig:mmst_dn_3d}}
    \quad
    \subfloat[3D style transfer on USA$\to$Sing.]{\includegraphics[width=0.3\linewidth]{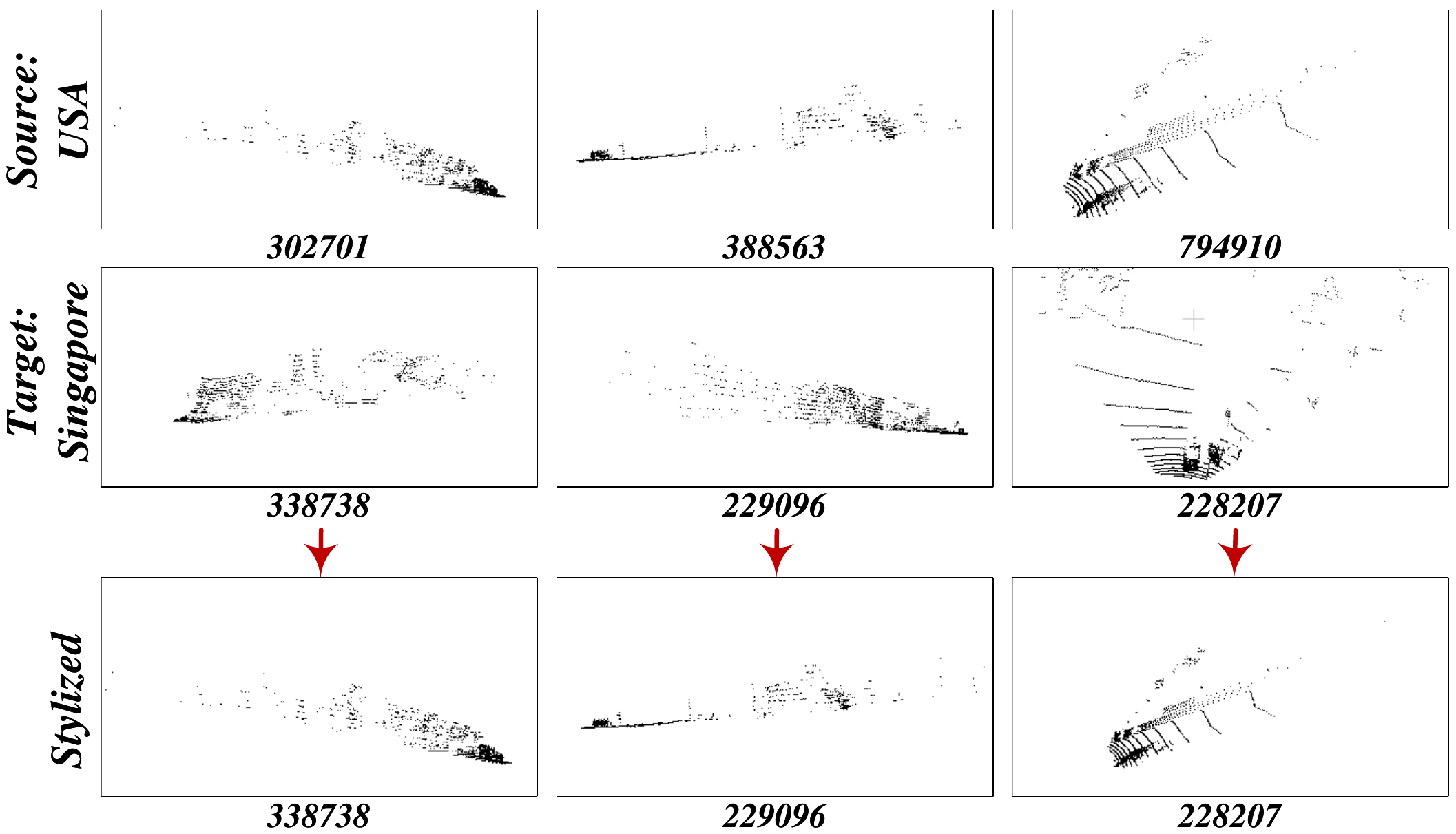}
    \label{fig:mmst_us_3d}}
    \quad
    \subfloat[3D style transfer on A2D2$\to$sKITTI]{\includegraphics[width=0.3\linewidth]{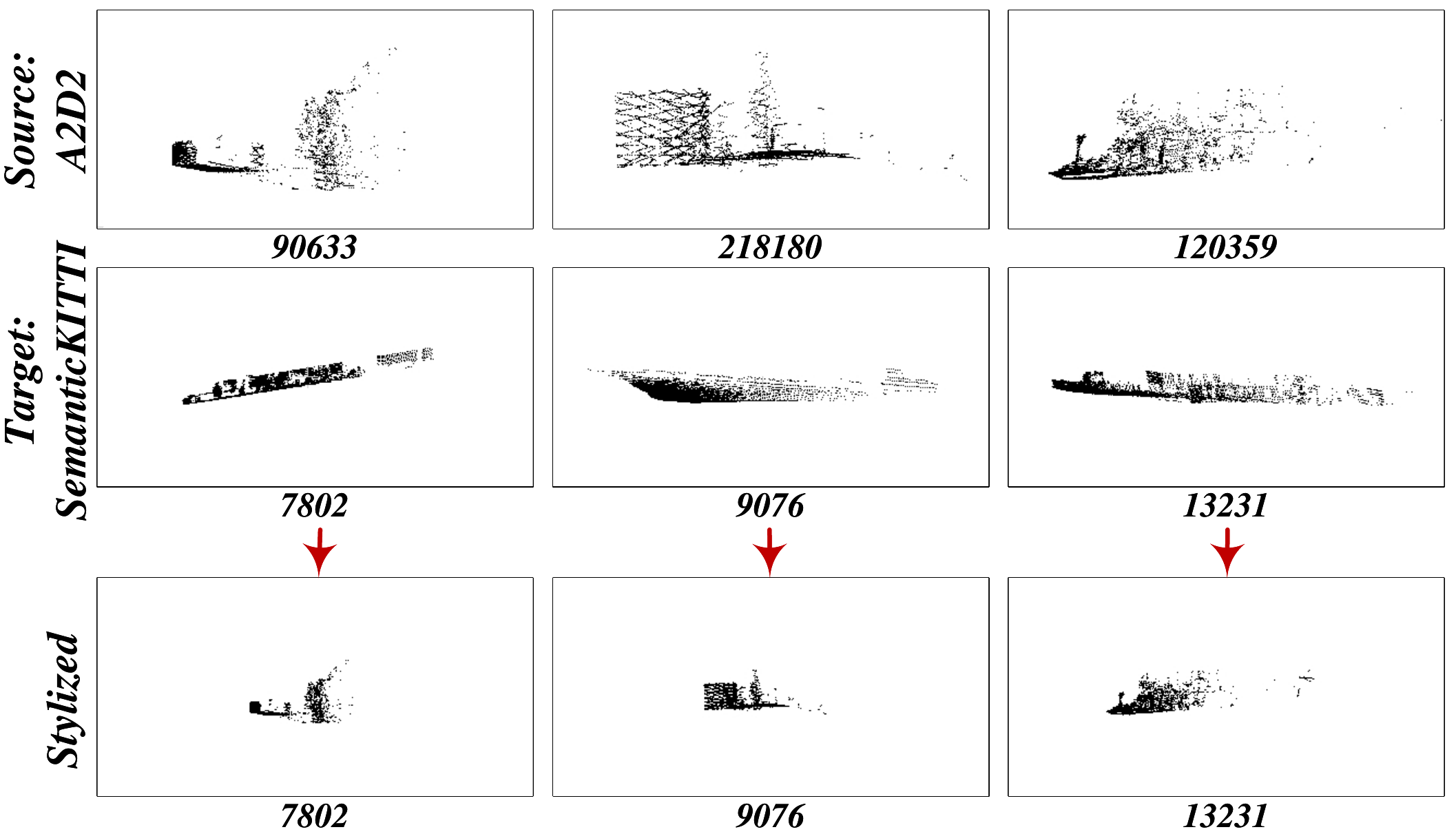}
    \label{fig:mmst_as_3d}}
    \caption{Style transfer on Day$\to$Night, USA$\to$Sing., and A2D2$\to$sKITTI adaptive scenarios. From top to bottom, they represent images and point clouds from the source, target, and style domains.}
\end{figure*}

\subsubsection{Effect of Stylized Input} To clearly show how the MMST module benefits the fusion-then-distillation strategy in our framework, we visualize the 2D and 3D style transfer results on some multi-modal scenarios. As shown in Fig.~\ref{fig:mmst_dn_2d}, Fig.~\ref{fig:mmst_us_2d}, and Fig.~\ref{fig:mmst_as_2d}, the source images in the first row are quite different from the target images in the second row. The last row shows the stylized images of the source domain in the target style. It is observed that stylized images are similar to the source images in terms of semantic content but relatively dark. Similarly, in Fig.~\ref{fig:mmst_dn_3d}, Fig.~\ref{fig:mmst_us_3d}, and Fig.~\ref{fig:mmst_as_3d}, stylized point clouds contain the same semantic content with source point clouds but sparser or denser according to the scan range of the target point clouds. In summary, these stylized images and point clouds form the augmented domain and are input into the \textit{Teacher} model to generate stylized features for cross-domain learning.

\subsubsection{Effect of Pseudo-Labeling}
\begin{figure}[t]
    \centering
    \includegraphics[width=1.0\linewidth]{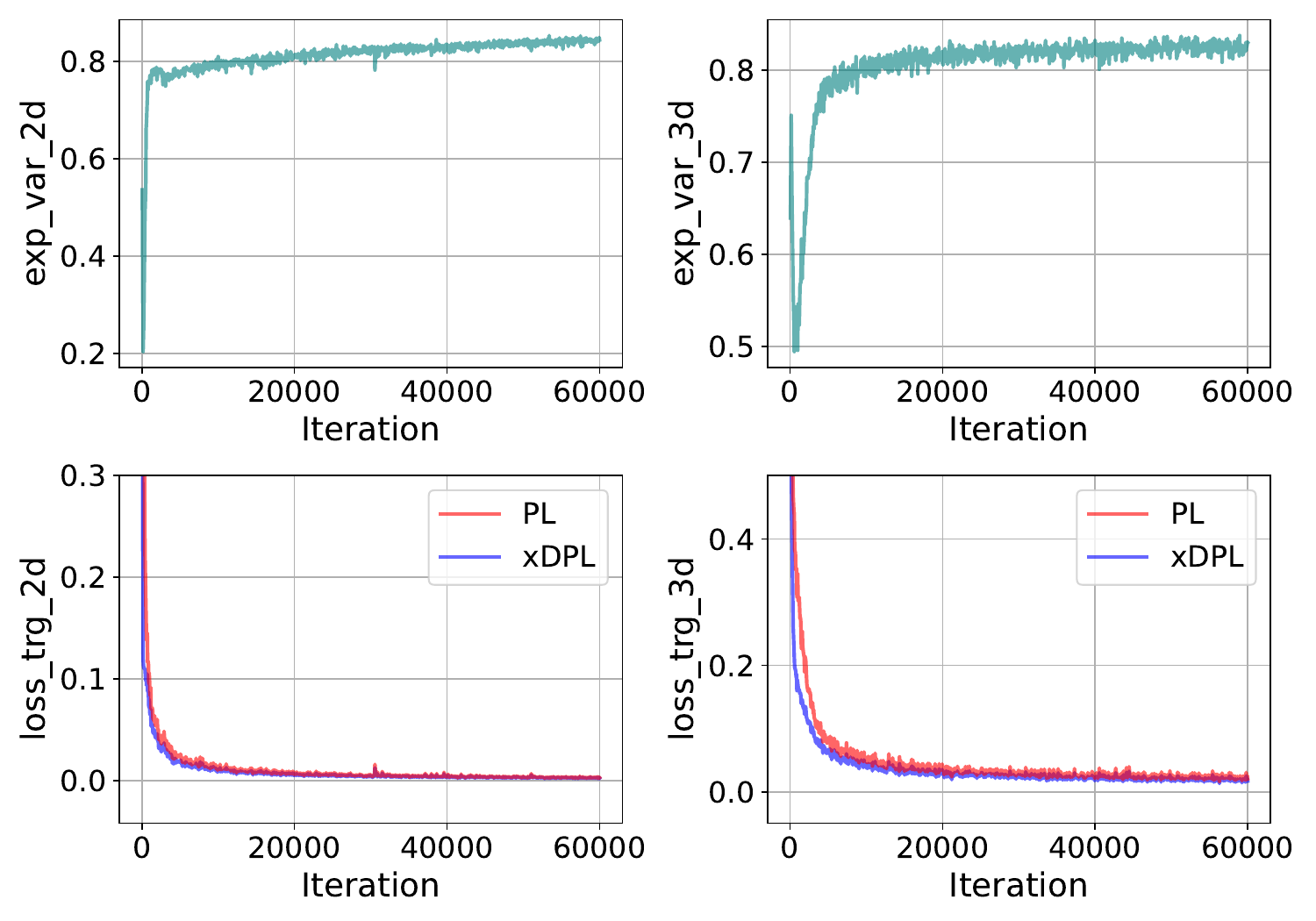}
    \caption{The variation curve of the self-training on USA$\to$Sing. scenario, where ``exp\_var\_2d'' and ``exp\_var\_3d'' denote the 2D and 3D exponential variances, respectively.}
    \label{fig:xdpl_loss}
\end{figure}

Self-training is a commonly used technique that typically employs pseudo-labels to learn from the unlabeled target data in UDA. However, after trial and error, we found that our model was unstable and easily degenerated when following the general pseudo-label in cross-modal UDA.
The main reason lies in one inherent problem the target pseudo-labels inevitably contain noise, which compromises the re-training of MFFM. Hence, we tabulate the results of our xDPL and common PL strategy in Tab.~\ref{tab:pl_xdpl}, which presents the effectiveness of xDPL in suppressing pseudo-label noise.
In Fig.~\ref{fig:xdpl_loss}, we further analyze why xDPL works. Firstly, the 1st and 2nd rows indicate the variations in multi-modal prediction variance and pseudo-label supervision loss in the target domain with training iterations.
It is observed that the exponential variance drops rapidly at the early stage of training, indicating that the prediction uncertainty of the model is high at this point.
Starting from 1k iterations, the variance shows a gradual upward trend, indicating that the reliability of the model predictions is increasing. By the time the model reaches 40k iterations, the predictions stabilize.
Due to the impact of prediction variance, the segmentation model focuses on learning reliable pixels and points, thereby achieving smaller 2D and 3D pseudo-label supervision losses.

\begin{table}[t]
  \centering
  \renewcommand{\arraystretch}{1.1}
  \caption{Ablation study of two pseudo-labeling strategies.}
  \resizebox{0.86\linewidth}{!}
  {
      \begin{tabular}{c|ccc|ccc}
        \toprule[1pt]
        \multirow{2}*{Method} & \multicolumn{3}{c|}{Day$\to$Night} & \multicolumn{3}{c}{USA$\to$Sing.} \\ \cline{2-7}
        ~ & 2D & 3D & xM & 2D & 3D & xM \\ \hline
        PL & 68.9 & 70.0 & 71.3 & 70.9 & 65.1 & 71.6 \\
        xDPL & 68.9 & 70.3 & 71.8 & 71.7 & 65.6 & 72.3 \\ \hline
        $\bigtriangleup_{gain}$ & $\uparrow$0.0 & \textcolor{teal}{$\uparrow$0.3} & \textcolor{teal}{$\uparrow$0.5} & \textcolor{teal}{$\uparrow$0.8} & \textcolor{teal}{$\uparrow$0.5} & \textcolor{teal}{$\uparrow$0.7} \\
        \bottomrule[1pt]
      \end{tabular}
    }
  \label{tab:pl_xdpl}
\end{table}
\begin{figure}[t]
    \centering
    \includegraphics[width=1.0\linewidth]{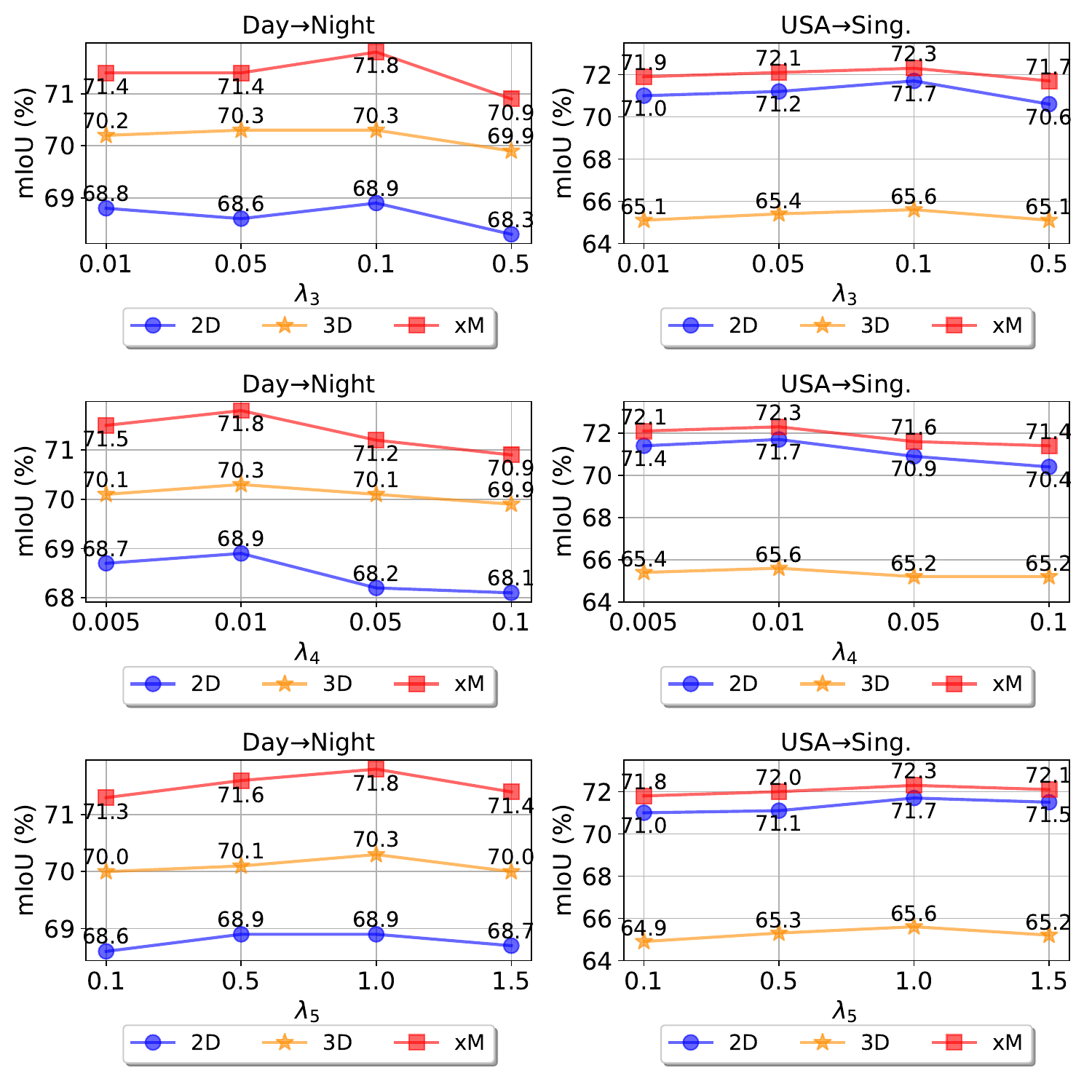}
    \caption{Parameter sensitivity analysis about the losses weights, including $\lambda_3, \lambda_4$ for DPD loss and $\lambda_5$ for xDPL loss.}
    \label{fig:lambda}
\end{figure}

\subsubsection{Parameter Sensitivity Analysis}
To testify whether our method is sensitive towards hyperparameter settings, we conduct parameter sensitivity analysis on Day$\to$Night and USA$\to$Sing. benchmarks with 2D, 3D, and xM metrics.
For the weights of MPD loss in each domain, we follow \cite{JaritzVCWP23} to fix $\lambda_1$ and $\lambda_2$ to 1.0 and 0.1, without any adjustments. But for the weights of DPD and xDPL losses, as shown in Fig.~\ref{fig:lambda}, we first fix $\lambda_{3}=0.1,\lambda_{5}=1.0$ and analyze the impact of $\lambda_4$, it is observed that when $\lambda_{4}=0.01$, the best results are achieved.
Similarly, we fix $\lambda_{4}=0.01,\lambda_{5}=1.0$ and analyze the impact of $\lambda_3$, it is observed that when $\lambda_{3}=0.1$, the best results are obtained.
Notably, although the fluctuation of $\lambda_3$ and $\lambda_4$ has a litter influence on bridging the domain gap, the fusion-then-distillation strategy is insensitive to hyperparameters compared to adversarial-based methods, indicating that our method is more stable.
Finally, we fix $\lambda_{3}=0.1,\lambda_{4}=0.01$, it is observed that the best $\lambda_5$ is set to $1.0$, where performance drops as the $\lambda_5$ becomes too large or too small.

\section{Conclusion}
In this paper, we investigate the non-negligible limitation aroused in domain adaptive 3D semantic segmentation (\textit{i.e., imbalanced modality adaptability}).
To solve this problem, we present FtD++ to explore cross-modal positive distillation of the source and target domains for 3D semantic segmentation.
FtD++ realizes distribution consistency between outputs not only for 2D images and 3D point clouds but also for source-domain and augment-domain and achieves state-of-the-art performance on several domain adaptive scenarios under unsupervised and semi-supervised settings.
We believe that our work will inspire a deeper investigation of cross-modal learning for domain adaptive 3D semantic segmentation in autonomous driving.



\bibliographystyle{IEEEtran}
\bibliography{main_arxiv}





\begin{IEEEbiography}[{\includegraphics[width=1in,height=1.25in,clip,keepaspectratio]{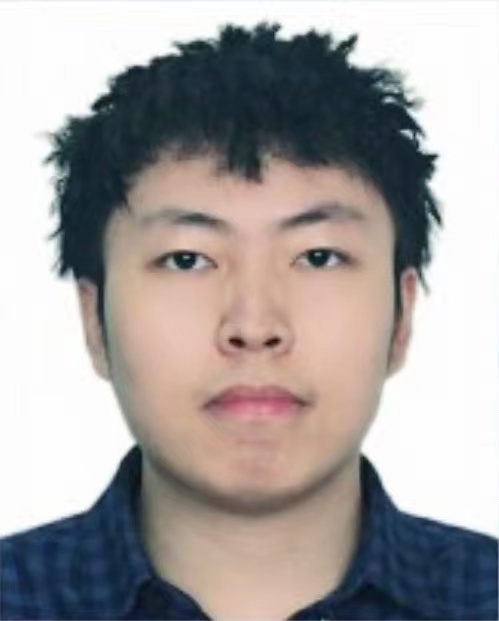}}]{Yao Wu} received the BE degree in the department of mechanical design manufacturing and automation, Fuzhou University, China, in 2017, and the ME degree in the department of mechanical engineering, Taiwan University, China, in 2019. He is currently working toward a PhD degree with the School of Informatics, Xiamen University, China. His research interests include machine learning and computer vision, including 3D scene understanding, visual-language models, domain adaptation, and semantic segmentation.
\end{IEEEbiography}


\begin{IEEEbiography}[{\includegraphics[width=1in,height=1.25in,clip,keepaspectratio]{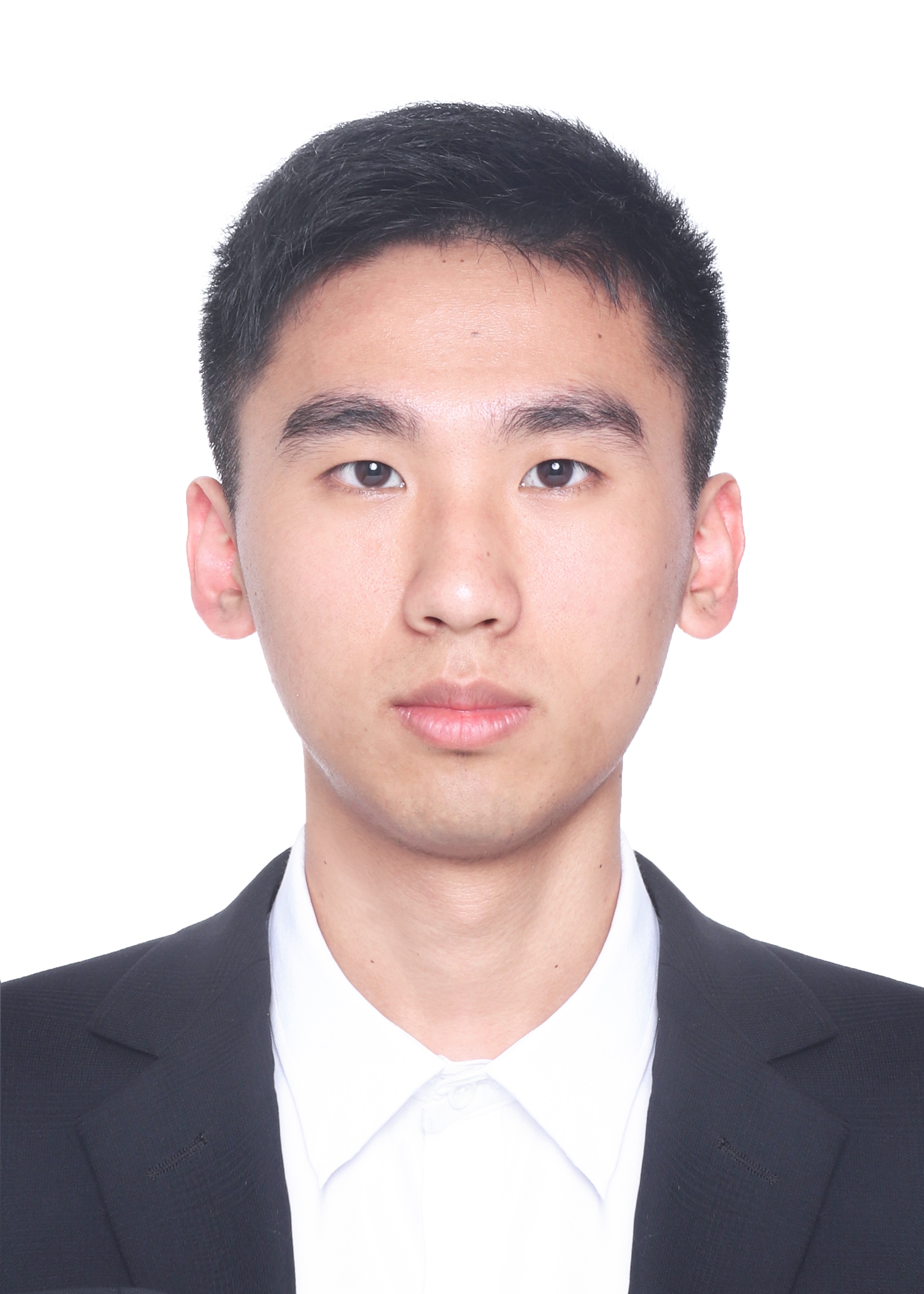}}]{Mingwei Xing} received the BE degree from the College of Computer and Data Science, Fuzhou University, China, in 2022. He is currently pursuing a Master's degree from the Institute of Artificial Intelligence, Xiamen University, China. His research interests include computer vision and semantic segmentation.
\end{IEEEbiography}


\begin{IEEEbiography}[{\includegraphics[width=1in,height=1.25in,clip,keepaspectratio]{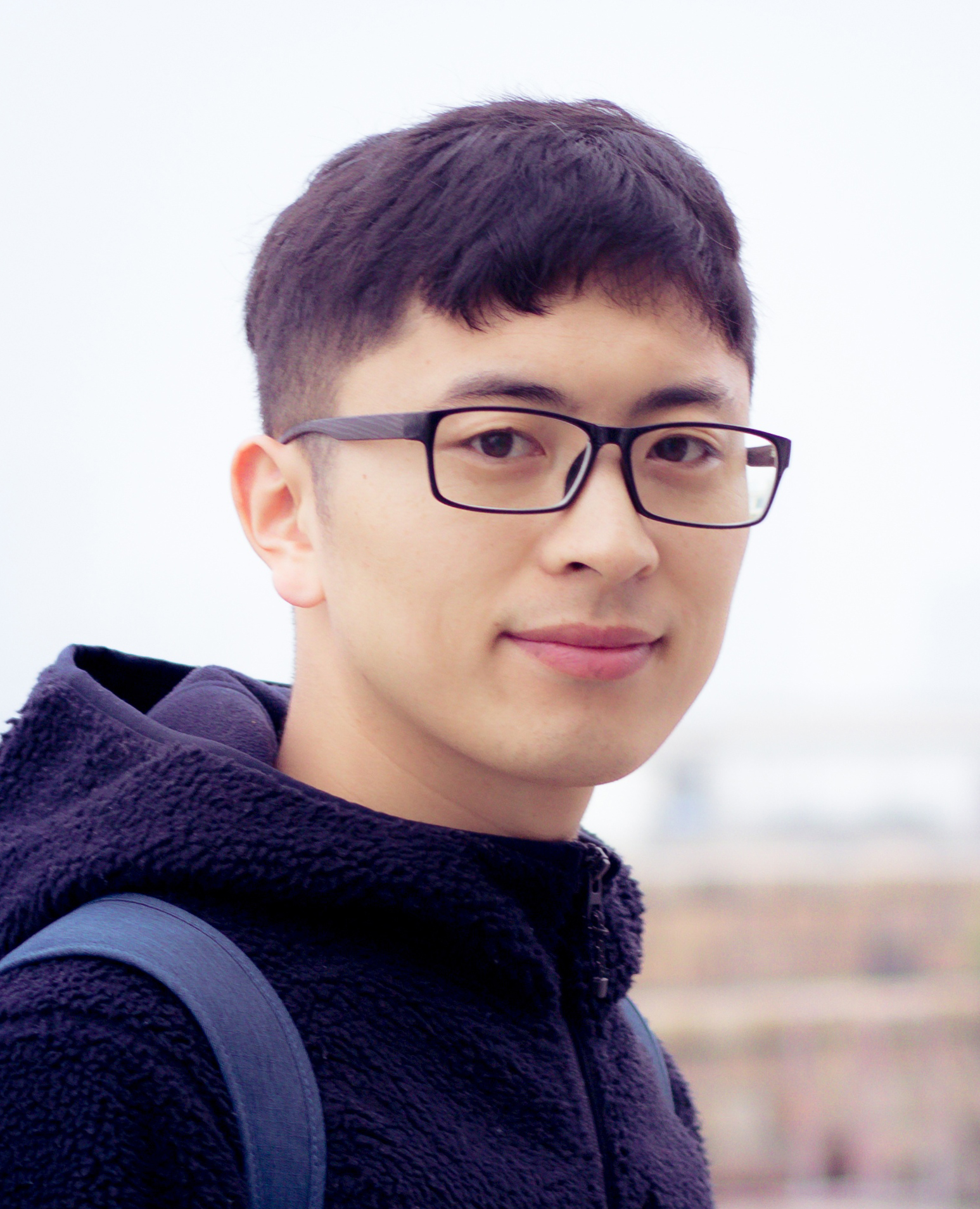}}]{Yachao Zhang} is a postdoctoral researcher at Intelligent Computing Lab, SIGS of Tsinghua University. He received the Ph.D. degree in Computer Science and Technology from Xiamen University, in 2022. His current research interests are machine learning and computer vision, including weakly-supervised learning, unsupervised learning, 3D scene understanding, point cloud semantic segmentation, etc. He has published 20 papers in major conferences and journals, including the TNNLS, AAAI, ICCV, ACMMM, CVPR, NeurIPS, IJCAI, etc. He also has served as a reviewer for more than 10 journals and conferences.
\end{IEEEbiography}


\begin{IEEEbiography}[{\includegraphics[width=1in,height=1.25in,clip,keepaspectratio]{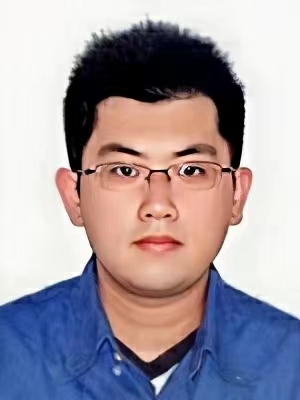}}]{Yuan Xie} (Member, IEEE) received the Ph.D. degree in pattern recognition and intelligent systems from the Institute of Automation, Chinese Academy of Sciences (CAS), Beijing, China, in 2013. He is currently a Full Professor with the School. of Computer Science and Technology, East China Normal University, Shanghai, China. He also has served as a reviewer for more than 15 journals and conferences. He has authored or coauthored around 85 papers in major international journals and conferences including the International Journal of Computer Vision (IJCV), the IEEE TRANSACTIONS ON PATTERN ANALYSIS AND MACHINE INTELLIGENCE (TPAMI), the IEEE TRANSACTIONS ON IMAGE PROCESSING (TIP), the IEEE TRANSACTIONS ON NEURAL NETWORKS AND LEARNING SYSTEMS (TNNLS), the IEEE TRANSACTIONS ON CYBERNETICS (TCYB), and the Conference and Workshop on Neural Information Processing Systems (NIPS), the International Conference on Machine Learning (ICML), the IEEE Conference on Computer Vision and Pattern Recognition (CVPR), the European Conference on Computer Vision (ECCV), and the IEEE International Conference on Computer Vision (ICCV). His research interests include image processing, computer vision, machine learning, and pattern recognition. Dr. Xie received the National Science Fund for Excellent Young Scholars of China in 2022.
\end{IEEEbiography}

\begin{IEEEbiography}[{\includegraphics[width=1in,height=1.25in,clip,keepaspectratio]{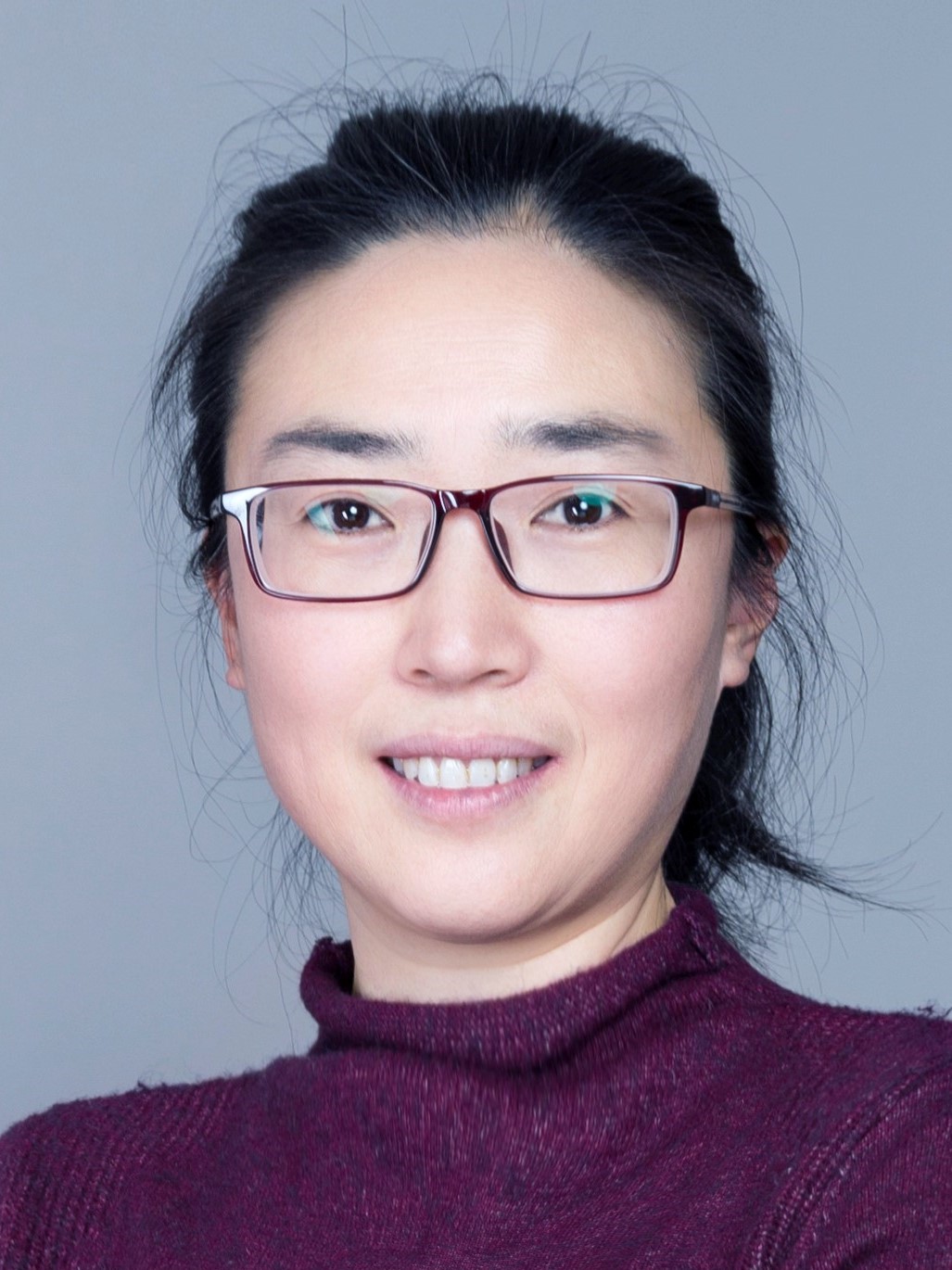}}]{Yanyun Qu} (Member, IEEE) received the Ph.D. degree in pattern recognition and intelligent systems from the Institute of Artificial Intelligence and Robotics, Xi’an Jiaotong University, Xi’an, China, in 2006. She is currently a Professor with the Department of Computer Science, School of Informatics, Xiamen University, Xiamen, China. She has authored and coauthored over 130 papers in major international journals and conferences, including the International Journal of Computer Vision, the IEEE TRANSACTIONS ON PATTERN ANALYSIS AND MACHINE INTELLIGENCE, the IEEE TRANSACTIONS ON IMAGE PROCESSING, the IEEE TRANSACTIONS ON CYBERNETICS, the IEEE TRANSACTIONS ON GEOSCIENCE AND REMOTE SENSING, Pattern Recognition, the IEEE International Conference on Computer Vision, the IEEE Conference on Computer Vision and Pattern Recognition, European Conference on Computer Vision, National Conference on Artificial Intelligence (AAAI), International Joint Conferences on Artificial Intelligence, Association for Computing Machinery (ACM) International Conference on Multimedia, and International Conference on Acoustics, Speech and Signal Processing. Her current research interests include image processing, computer vision, machine learning, and pattern recognition. Dr. Qu is a member of ACM and the Secretary of the Technical Committee of Hybrid Artificial Intelligence and Chinese Association of Automation.
\end{IEEEbiography}

\vfill

\end{document}